\documentclass{article}

\usepackage{microtype}
\usepackage{graphicx}
\usepackage{subcaption}
\usepackage{booktabs} 
\usepackage{tcolorbox}
\usepackage{xcolor}
\usepackage[dvipsnames]{xcolor}
\usepackage{makecell}

\usepackage{hyperref}

\usepackage[accepted]{icml2026}

\usepackage{amsmath}
\usepackage{amssymb}
\usepackage{mathtools}
\usepackage{amsthm}
\usepackage{stmaryrd}

\usepackage[capitalize,noabbrev]{cleveref}

\usepackage[textsize=tiny,textwidth=1.15\linewidth]{todonotes}
\makeatletter
\tikzset{
    notestyleraw/.append style={
        fill opacity=0.8,    
        text opacity=1       
    }
}
\makeatother
\usepackage{amsthm}
\usepackage{amssymb}
\usepackage{algorithm}
\usepackage{amsmath}
\usepackage{cite}
\usepackage{acronym}

\usepackage{commath,amsmath,amssymb,amsfonts}

\usepackage{accents}
\usepackage[inkscapearea=page]{svg}

\usepackage{mathtools}
\usepackage{hyperref}
\usepackage{amssymb}
\usepackage{bm}
\usepackage{graphicx}
\usepackage{float}
\usepackage{amsmath}
\usepackage{tikz}
\usepackage{multicol}
\usepackage{dsfont}
\usepackage{tcolorbox}
\usepackage{bbm}
\usepackage{multirow}
\usetikzlibrary{positioning, backgrounds, fit, shapes.arrows, shapes.geometric, matrix}
\usetikzlibrary{decorations.markings}

\usepackage{enumitem}
\usepackage{caption}
\usepackage{booktabs}
\usepackage{subcaption}
\tikzset{block/.style = {draw, fill=white, rectangle,
		minimum height=3em, minimum width=2cm},
	input/.style = {coordinate},
	output/.style = {coordinate},
	pinstyle/.style = {pin edge={to-,t,black}}
	radiation/.style={{decorate,decoration={expanding waves,angle=90,segment   length=4pt}}}
	
}
\usepackage{smartdiagram}

\tikzstyle{block} = [draw, rectangle, minimum height=2em, minimum width=2em]
\tikzstyle{sum} = [draw, circle,minimum width=0.1 cm]
\tikzstyle{input} = [coordinate]
\tikzstyle{output} = [coordinate]
\tikzstyle{dummy} = [coordinate]
\tikzstyle{pinstyle} = [pin edge={to-,thin,black}]
\usetikzlibrary{positioning, fit, arrows.meta}
\usetikzlibrary{positioning}
\usetikzlibrary{shapes,arrows}
\tikzstyle{frame_cyan} = [thick, draw=blue, solid,inner sep=0.3em]
\tikzstyle{frame_red} = [thick, draw=red, solid,inner sep=0.3em]
\tikzstyle{frame_green} = [thick, draw=green, solid,inner sep=0.3em]

\usepackage{tikz}
\usepackage{xcolor}
\definecolor{fc}{HTML}{1E90FF}
\definecolor{h}{HTML}{228B22}
\definecolor{bias}{HTML}{87CEFA}
\definecolor{noise}{HTML}{8B008B}
\definecolor{conv}{HTML}{FFA500}
\definecolor{pool}{HTML}{B22222}
\definecolor{up}{HTML}{B22222}
\definecolor{view}{HTML}{FFFFFF}
\definecolor{bn}{HTML}{FFD700}
\tikzset{fc/.style={black,draw=black,fill=fc,rectangle,minimum height=1cm}}
\tikzset{h/.style={black,draw=black,fill=h,rectangle,minimum height=1cm}}
\tikzset{bias/.style={black,draw=black,fill=bias,rectangle,minimum height=1cm}}
\tikzset{noise/.style={black,draw=black,fill=noise,rectangle,minimum height=1cm}}
\tikzset{conv/.style={black,draw=black,fill=conv,rectangle,minimum height=1cm}}
\tikzset{pool/.style={black,draw=black,fill=pool,rectangle,minimum height=1cm}}
\tikzset{up/.style={black,draw=black,fill=up,rectangle,minimum height=1cm}}
\tikzset{view/.style={black,draw=black,fill=view,rectangle,minimum height=1cm}}
\tikzset{bn/.style={black,draw=black,fill=bn,rectangle,minimum height=1cm}}
 
\usepackage{xspace}

\tikzstyle{dummy} = [coordinate]
\pgfkeys{/pgf/.cd,
  parallelepiped offset x/.initial=2mm,
  parallelepiped offset y/.initial=2mm
}
\pgfdeclareshape{parallelepiped}
{
  \inheritsavedanchors[from=rectangle] 
  \inheritanchorborder[from=rectangle]
  \inheritanchor[from=rectangle]{north}
  \inheritanchor[from=rectangle]{north west}
  \inheritanchor[from=rectangle]{north east}
  \inheritanchor[from=rectangle]{center}
  \inheritanchor[from=rectangle]{west}
  \inheritanchor[from=rectangle]{east}
  \inheritanchor[from=rectangle]{mid}
  \inheritanchor[from=rectangle]{mid west}
  \inheritanchor[from=rectangle]{mid east}
  \inheritanchor[from=rectangle]{base}
  \inheritanchor[from=rectangle]{base west}
  \inheritanchor[from=rectangle]{base east}
  \inheritanchor[from=rectangle]{south}
  \inheritanchor[from=rectangle]{south west}
  \inheritanchor[from=rectangle]{south east}
  \backgroundpath{
    \southwest \pgf@xa=\pgf@x \pgf@ya=\pgf@y
    \northeast \pgf@xb=\pgf@x \pgf@yb=\pgf@y
    \pgfmathsetlength\pgfutil@tempdima{\pgfkeysvalueof{/pgf/parallelepiped offset x}}
    \pgfmathsetlength\pgfutil@tempdimb{\pgfkeysvalueof{/pgf/parallelepiped offset y}}
    \def\ppd@offset{\pgfpoint{\pgfutil@tempdima}{\pgfutil@tempdimb}}
    \pgfpathmoveto{\pgfqpoint{\pgf@xa}{\pgf@ya}}
    \pgfpathlineto{\pgfqpoint{\pgf@xb}{\pgf@ya}}
    \pgfpathlineto{\pgfqpoint{\pgf@xb}{\pgf@yb}}
    \pgfpathlineto{\pgfqpoint{\pgf@xa}{\pgf@yb}}
    \pgfpathclose
    \pgfpathmoveto{\pgfqpoint{\pgf@xb}{\pgf@ya}}
    \pgfpathlineto{\pgfpointadd{\pgfpoint{\pgf@xb}{\pgf@ya}}{\ppd@offset}}
    \pgfpathlineto{\pgfpointadd{\pgfpoint{\pgf@xb}{\pgf@yb}}{\ppd@offset}}
    \pgfpathlineto{\pgfpointadd{\pgfpoint{\pgf@xa}{\pgf@yb}}{\ppd@offset}}
    \pgfpathlineto{\pgfqpoint{\pgf@xa}{\pgf@yb}}
    \pgfpathmoveto{\pgfqpoint{\pgf@xb}{\pgf@yb}}
    \pgfpathlineto{\pgfpointadd{\pgfpoint{\pgf@xb}{\pgf@yb}}{\ppd@offset}}
  }
}
\pgfdeclareshape{document}{
\inheritsavedanchors[from=rectangle] 
\inheritanchorborder[from=rectangle]
\inheritanchor[from=rectangle]{center}
\inheritanchor[from=rectangle]{north}
\inheritanchor[from=rectangle]{north east}
\inheritanchor[from=rectangle]{north west}
\inheritanchor[from=rectangle]{south}
\inheritanchor[from=rectangle]{south east}
\inheritanchor[from=rectangle]{south west}
\inheritanchor[from=rectangle]{west}
\inheritanchor[from=rectangle]{east}
\backgroundpath{%
\southwest \pgf@xa=\pgf@x \pgf@ya=\pgf@y
\northeast \pgf@xb=\pgf@x \pgf@yb=\pgf@y
\pgf@xc=\pgf@xb \advance\pgf@xc by-5pt 
\pgf@yc=\pgf@ya \advance\pgf@yc by5pt
\pgfpathmoveto{\pgfpoint{\pgf@xa}{\pgf@ya}}
\pgfpathlineto{\pgfpoint{\pgf@xa}{\pgf@yb}}
\pgfpathlineto{\pgfpoint{\pgf@xb}{\pgf@yb}}
\pgfpathlineto{\pgfpoint{\pgf@xb}{\pgf@yc}}
\pgfpathlineto{\pgfpoint{\pgf@xc}{\pgf@ya}}
\pgfpathclose
\pgfpathmoveto{\pgfpoint{\pgf@xc}{\pgf@ya}}
\pgfpathlineto{\pgfpoint{\pgf@xc}{\pgf@yc}}
\pgfpathlineto{\pgfpoint{\pgf@xb}{\pgf@yc}}
\pgfpathclose
}
}
\tikzstyle{block} = [draw, fill=white, rectangle, 
    minimum height=3em, minimum width=6em]
    
\usetikzlibrary{backgrounds}
\usepackage{pifont}
\tikzstyle{startstop} = [rectangle, rounded corners, minimum width=2cm, minimum height=0.7cm,text centered, draw=black, fill=lime!30]
\usepackage{stackengine}

\makeatletter
\newcounter{phase}[algorithm]
\newlength{\phaserulewidth}
\newcommand{\setphaserulewidth}{\setlength{\phaserulewidth}}

\makeatother
\setphaserulewidth{.7pt}
\makeatother
\usepackage{colortbl} 
\newcommand\MakeUppercaseGreek[1]{
  \begingroup
    \let\psi\Psi
    \let\omega\Omega
    \let\gamma\Gamma
    \MakeUppercase{#1}
  \endgroup}

\newcommand{\matsym}[1]{\mathbf{#1}}

\newcommand{\diag}[1]{\mathrm{Diag}\left(#1\right)}

\newcommand{\probP}{\text{I\kern-0.15em P}}

\def\rmd{\mathrm{d}}

\newcommand{\intset}[2]{\llbracket #1, #2 \rrbracket}

\newcommand*\circled[1]{\tikz[baseline=(char.base)]{%
            \node[shape=circle,fill=black!10!white,inner sep=1.5pt] (char) {\textbf{#1}};}}

\def\idw{{\sc{IDW}}}
\def\gmz{{\sc{GMZ}}}
\def\ok{{\sc{OK}}}
\def\RM{{\sc{RM}}}

\def\mgdm{{\sc{MGDM}}}
\def\dps{{\sc{DPS}}}
\def\tds{{\sc{TDS}}}
\def\crepe{{\sc{CREPE}}}
\def\mgps{{\sc{MGPS}}}
\def\reddiff{{\sc{RedDiff}}}
\def\daps{{\sc{DAPS}}}

\def\rmse{{RMSE}}

\def\pcc{{PCC}}

\newcommand{\Id}{\matsym{I}}

\definecolor{BaseHighlight}{HTML}{1f77b4}
\newcommand{\first}[1]{\colorbox{BaseHighlight!50}{#1}}
\newcommand{\second}[1]{\colorbox{BaseHighlight!20}{#1}}

\def\codeURL{https://github.com/Badr-MOUFAD/rainfield-diffusion-models}
\def\codeURLShort{github.com/Badr-MOUFAD/rainfield-diffusion-models}

\renewcommand{\cite}[1]{\citep{#1}}

\usepackage{amsmath,amssymb}
\usepackage{float}

\begin{document}

\twocolumn[
  \icmltitle{Bayesian Rain Field Reconstruction using Commercial Microwave Links and Diffusion Model Priors}



  \icmlsetsymbol{equal}{*}

  \begin{icmlauthorlist}
    \icmlauthor{Badr Moufad}{equal,cmap}
    \icmlauthor{Albina Ilina}{equal,mbz}
    \icmlauthor{Hai Victor Habi}{tel}
    \icmlauthor{Salem Lahlou}{mbz}
    \icmlauthor{Yazid Janati}{ifm}
    \icmlauthor{Hagit Messer}{tel}
    \icmlauthor{Eric Moulines}{mbz,epita}
  \end{icmlauthorlist}

  \icmlaffiliation{cmap}{CMAP, Ecole Polytechnique, France}
  \icmlaffiliation{mbz}{MBZUAI, UAE}
  \icmlaffiliation{ifm}{IFM, France}
  \icmlaffiliation{epita}{EPITA, France}
  \icmlaffiliation{tel}{School of Electrical and Computer Engineering, Tel Aviv University, Tel Aviv, Israel}

  \icmlcorrespondingauthor{Badr Moufad}{badr.moufad@polytechnique.edu}
  \icmlcorrespondingauthor{Hai Victor Habi}{haivictorh@mail.tau.ac.il}

  \icmlkeywords{Machine Learning, ICML}

  \vskip 0.3in
]



\printAffiliationsAndNotice{\icmlEqualContribution}

\begin{abstract}
  Commercial Microwave Links (CMLs) offer dense spatial coverage for rainfall sensing but produce path-integrated measurements that make accurate ground-level reconstruction challenging.
  Existing methods typically oversimplify CMLs as point sensors and neglect line integration relating rainfall to signal attenuation, resulting in degraded performance under heterogeneous precipitation.
  In this work, we view rain field reconstruction as a Bayesian inverse problem with Diffusion Models (DMs) as high-fidelity spatial priors.
  We show that diffusion models better preserve key rainfall statistics compared to censored Gaussian processes.
  Framing rainfall estimation as a Bayesian inverse problem with a DM prior enables training-free posterior sampling using a broad family of methods, including Plug-and-Play, Sequential Monte Carlo, and Replica Exchange methods.
  Experiments on synthetic and real-world datasets demonstrate consistent improvements over established CML-based reconstruction baselines.
\end{abstract}

\section{Introduction}
Accurate rain fields are central to hydrology, flood warning, and water resource management. Standard sensing relies on Rain Gauges (RGs), weather radars, and satellites.
RGs are accurate but sparse \citep{messer2022rain}.
Weather radars provide broad coverage, yet require calibration and are sensitive to clutter and atmospheric effects \citep{michelson2000gaugeradar,krajewski2002radar,harrison2000improving}.
Weather satellites extend coverage to data-scarce regions but often trade resolution for reach \citep{Funk2014CHIRPS}.
Hence, no single modality provides dense, accurate, and affordable precipitation monitoring at scale.
Opportunistic remote sensing addresses this gap by reusing existing infrastructure.
In particular, \emph{Commercial Microwave Links} (CMLs) in cellular backhaul networks can act as near-ground rain sensors \citep{messer2006environmental,uijlenhoet2018opportunistic,graf2020rainfall,graf2026opportunistic}. CMLs form dense networks of transmitters and receivers deployed for telecommunication purposes. During precipitation, rainfall induces attenuation along each link path $\{\mathcal L_i\}_{i=1}^m$ of the network, which is commonly modeled by the empirical power law \citep{leijnse2007rainfall,eshel2021quantitative}
\begin{equation}
\label{eq:power-law}
Y_i
= a \int_{\mathbf{s}\in\mathcal{L}_i} X(\mathbf{s})^b \,\rmd \mathcal{L}_i \;+\; \sigma_i Z_i,
\quad Z_i\sim\mathcal N(0,1),
\end{equation}
where $X(\mathbf{s})$ is the rain rate (mm/h), and $(a,b)$ are constants that depend on the characteristics of the link, while $\sigma_i Z_i$ is an additive Gaussian noise of standard deviation $\sigma_i$ independent of the attenuation $Y_i$.

Rain-field reconstruction from CMLs is an inverse problem: for a region of interest $\Omega \subset \mathbb{R}^2$, and given attenuations $\{Y_i\}_{i=1}^m$, estimate the gridded field in an $H \times W$ domain.
The problem is ill-posed because measurements are path-integrated and nonlinear.
Many reconstruction methods approximate each link as a \emph{Virtual Rain Gauge} (VRG) and then interpolate \citep{shepard1968two,overeem2013country,goldshtein2009idw-gmz,eshel2021quantitative}, which can discard valuable spatial information and degrade under heterogeneous rainfall. More elaborate variants, e.g., multiple VRGs per link, still inherit interpolation limitations \citep{goldshtein2009idw-gmz}. The fundamental challenge is that CML measurements provide aggregate information along paths, whereas reconstruction requires resolving fine-scale spatial structure. This mismatch is a hallmark of ill-posed inverse problems: infinitely many rain fields can produce identical observations.
Regularization through prior knowledge is essential, yet classical choices such as isotropic smoothness or Gaussian spatial models fail to capture the complex, multi-scale, and intermittent structure of real precipitation.
In this work, we instead cast CML-based rain field reconstruction as Bayesian inference with a learned generative prior. Recent works show that deep generative models, especially \emph{Diffusion Models} (DMs), provide strong priors for inverse problems and are amenable to \emph{training-free} posterior sampling via inference-time guidance \citep{daras2024dm-survey,janati2025dm-survey,chung2023dps}.
This setting consists in training a DM offline on representative data and then reusing it at inference as a prior to regularize ill-posed inverse problems.

Building on this paradigm, we make the following contributions.
\circled{1} We show that DM priors yield high-fidelity spatial rain fields and improve key statistics, e.g. cumulative precipitation, compared to censored Gaussian Processes.
\circled{2} We leverage training-free posterior sampling to enable flexible inference with a fixed pre-trained DM, covering Plug-and-Play, Sequential Monte Carlo, and Replica Exchange approaches.
\circled{3} We move beyond VRG approximations by explicitly modeling the nonlinear, path-integrated CML operator governed by the empirical the power-law.

We validate the approach on simulated and real datasets, and observe consistent gains over established baselines.
To the best of our knowledge, this is the first work to bring posterior sampling with diffusion priors to opportunistic rain sensing with CMLs.
We also release our code%
\footnote{Link: \small{\href{\codeURL}{\codeURLShort}}}.

\section{Related Work}
\label{sec:related work}

CMLs were originally proposed as opportunistic rainfall sensors based on the physically grounded relationship between rain-induced signal attenuation and path-integrated rainfall intensity \citep{messer2006environmental}.
A dominant modeling assumption in much of the CML literature is to reduce each path-integrated CML measurement to a single point observation, typically assigned to the link midpoint, thereby ignoring rainfall variability along the link path.
Under this abstraction, rainfall fields are reconstructed using deterministic point-based spatial interpolation methods such as Inverse Distance Weighting (\idw) and Ordinary Kriging (\ok) \citep{zhang2023reconstructing, blettner2022combining, messer2022rain}.
\idw\ estimates values at grid locations as weighted averages of nearby measurements, with weights decaying as a power of the distance, whereas \ok\ exploits second-order spatial statistics and derives interpolation weights from the spatial correlation structure of the observed data through a variogram model.
While these approaches are simple and computationally efficient, collapsing line-integral measurements into point estimates inherently discards rainfall variability information contained in the line-integral nature of CML measurements along the link path. As a consequence, the reconstructed fields are often overly smooth and fail to reproduce fine-scale spatial variability and localized extremes.
To mitigate this limitation, \citet{messer2022rain,eshel2021quantitative} proposed Goldshtein-Messer-Zinevich (\gmz)
algorithm where multiple VRGs are used to approximate the path-integrated measurement.
The VRGs are placed along the link, and their intensities are iteratively adjusted to ensure consistency with the original CML observation. While GMZ better represents sub-link variability, it ultimately remains within the interpolation framework, as it relies on pseudo-point measurements rather than directly assimilating the path-integral observations.

More advanced stochastic reconstruction methods \cite{haese2017stochastic,horning2019random-mising,blettner2022combining}, such as Random Mixing (\RM), account for the path-integrated nature of CML measurements and, unlike interpolation methods, provide a principled treatment of reconstruction uncertainty. It generates precipitation fields as stochastic linear combinations of unconditional random fields that reproduce the observed spatial dependence structure. RG measurements are imposed as pointwise constraints, while CML observations enter as nonlinear, path-integrated constraints. Repeating this procedure over several steps yields an ensemble of spatial fields, allowing reconstruction uncertainty to be quantified. 
However, RM relies on the generation and optimization of large ensembles of unconditional random fields, which is computationally expensive, especially for country-scale applications. Besides, it is constrained by the need to specify marginal rainfall distributions and spatial dependence structures. This makes them rain-gauge dependent, as the reliable estimation of these statistical properties typically requires a sufficiently dense and gauge network for the marginals. Consequently, such methods are not well suited for use cases in which only a very limited number of rain gauges are available, or where rainfall information is derived almost exclusively from CMLs. 

In this work, we focus on rain field reconstruction using only CML measurements.
We provide an extended discussion of deterministic and stochastic  CML-based reconstruction methods, including the Random Mixing framework and multi-sensor fusion approaches, in \Cref{sec:extended_related_work}.


\section{Rain Field Prior Models}
\begin{figure}[t]
    \centering%
    \includegraphics[width=0.9\linewidth]{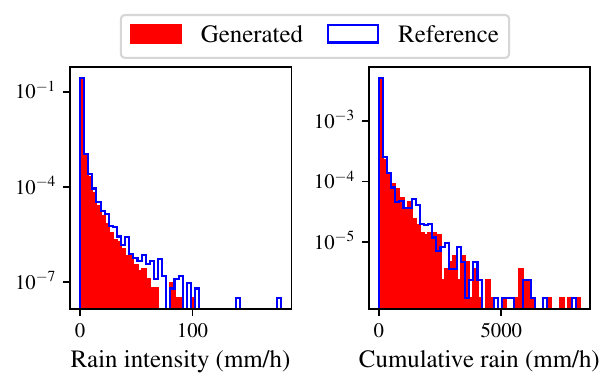}%
    \caption{Comparison of rain field statistics between the reference samples and generated samples using the DM prior. The histograms show densities and are built using $5000$ samples.}
    \label{fig:stats-dm-prior}
\end{figure}

\begin{figure*}[t]
    \centering
    \hspace*{10mm}%
    \begin{subfigure}[t]{0.4\textwidth}
        \centering
        \includegraphics[width=\textwidth]{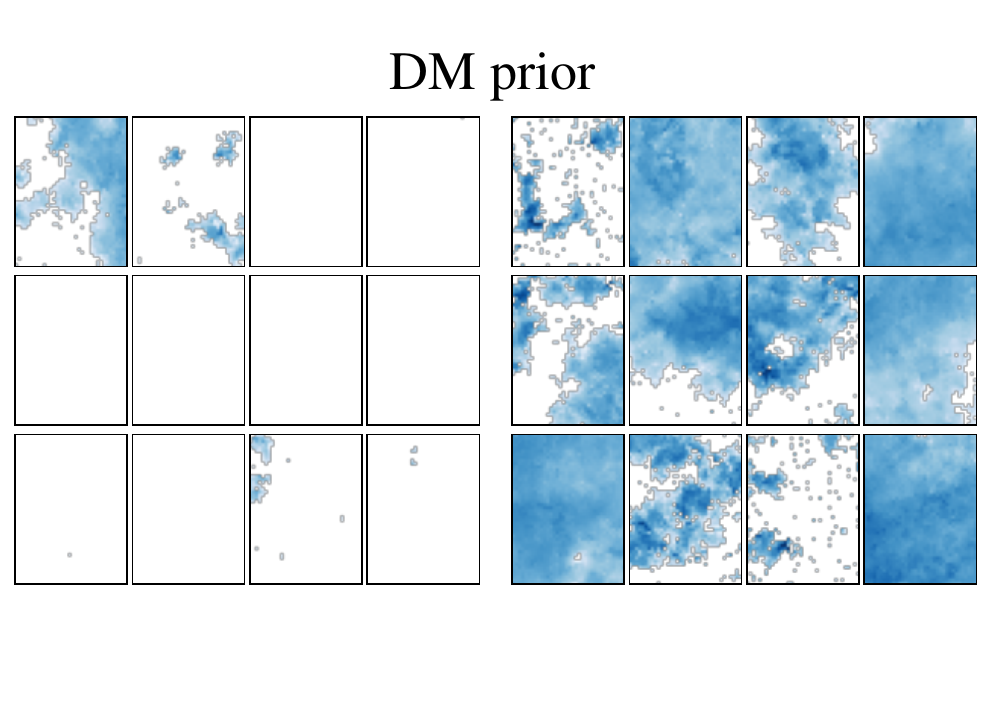}
    \end{subfigure}%
    \hspace*{5mm}%
    \begin{subfigure}[t]{0.4\textwidth}
        \centering
        \includegraphics[width=\textwidth]{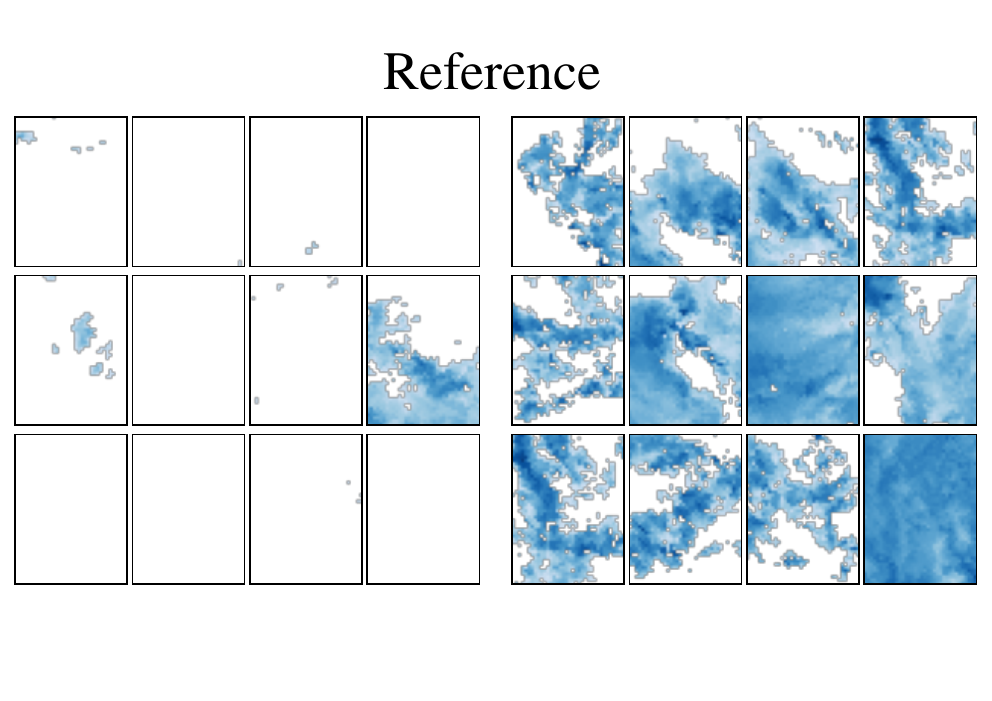}
    \end{subfigure}%
    \hfill%
    \begin{subfigure}[t]{0.08\textwidth}
        \centering
        \vspace{-37mm} 
        \includegraphics[width=1.\textwidth]{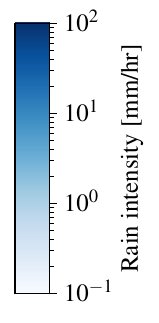}
    \end{subfigure}%
    \captionsetup{font=small}
    \caption{Examples of generated rain fields using the prior DM compared with reference rain fields.
    A total of $5{,}000$ samples were generated.
    In each subfigure, on the left, samples are randomly drawn from the datasets; on the right, they are drawn from the top $90\%$ wet samples by cumulative rain.}
    \label{fig:generated-samples}
\end{figure*}

The choice of prior is critical for regularizing the ill-posed CML inverse problem. We compare two modeling approaches: (i) a censored Gaussian process that explicitly encodes rainfall characteristics such as spatial correlation and non-negativity, and (ii) DM that learns them directly from data.

\subsection{Censored Gaussian Processes}
\label{sec:censored-gp}
Rainfall intensity fields are inherently intermittent: they combine frequent zero-precipitation events with highly skewed positive intensities, as illustrated in \Cref{fig:stats-dm-prior}. There exists a rich literature on stochastic precipitation modeling and geostatistical reconstruction \citep{wilks1999weather,wheater2000spatial,yang2005spatial}. A classical and widely used statistical abstraction models this mixed behavior through a latent Gaussian Process that is left censored at zero \citep{bardossy1992}. Early frameworks \citep{bardossy1992,ailliot2009} represent spatial dependence by a latent Gaussian random field and enforce non-negativity by censoring, often together with power-type transformations to better match the heavy right tail of positive rain rates. Subsequent work refined these ideas by introducing more complex transformations of the censored latent process in order to increase flexibility while retaining analytic and computational convenience \cite{baxevani2015,stauffer2017}. 
Despite their practical appeal, censored Gaussian Process (GP) priors remain limited by the structure of Gaussian kernels. In particular, Gaussian-based kernels tend to induce comparatively \emph{stiff} priors: they can struggle to reproduce the sharp gradients, multi-scale variability, and nonstationary spatial organization that commonly arise in rain fields. In addition, when these models are used as priors within Bayesian inverse problems, inference can become computationally demanding. The censoring operation introduces inequality constraints and requires manipulating high-dimensional truncated or censored Gaussian distributions, which in turn involves prohibitive sampling procedures \citep{pakman2014exact,botev2017normal}.

We adopt the following censored GP model as a classical baseline prior. The rain field $X$ is defined on a $H \times W$ grid and is constructed from a latent Gaussian random field $V$. Specifically, we assume
$V \sim \mathcal{GP}(\mu, k)$ with $\mu: \mathbb{R}^2 \rightarrow \mathbb{R}$.
To accommodate anisotropy, we use a stationary kernel parameterized by a positive definite matrix $\mathbf{Q}\in\mathbb{R}^{2\times 2}$:
\begin{equation*}
    k(\mathbf{s}, \mathbf{s}') = \sigma^2 \exp \left( -\frac{1}{2} ( \mathbf{s} - \mathbf{s}' )^\top \mathbf{Q} \ ( \mathbf{s} - \mathbf{s}' ) \right),
\end{equation*}
where $\mathbf{s}, \mathbf{s}' \in \mathbb{R}^2$. The observed rain field is then obtained via a power transformation followed by censoring at zero,
$$
X(\mathbf{s}) = \max\big(0, V(\mathbf{s})^\beta\big),
\quad \mathrm{for} \;\; \mathbf{s} \in \mathbb{R}^2,
$$
which induces a point mass at zero together with a skewed positive component. The parameter $\beta$ controls the degree of skewness in the positive intensities, while $\mathbf{Q}$ and $\sigma$ govern the spatial correlation structure and marginal scale, respectively. This model is parameterized by $(\mu, \mathbf{Q}, \sigma, \beta)$.
The Expectation--Maximization algorithm \citep{ordonez2018geostatistical} is used to estimate the parameters of the model $(\mu, \mathbf{Q}, \sigma, \beta)$ which treats the latent field $V$ as missing data and accounts for the censoring mechanism during inference; details are provided in \Cref{apdx:censored-gp}.

\subsection{Diffusion Models}
Let $x_0 \sim p_0$ denote a sample from the data distribution. We adopt the \emph{Variance Exploding} (VE) formulation of Diffusion Models defined by fixing an increasing noise schedule
$0 = \sigma_0 < \sigma_1 < \cdots < \sigma_T$ \citep{song2021dm-sde,karras2022edm}.
The corruption process is defined as a Markov chain that gradually perturbs the data via Gaussian noise,
\begin{equation}
\label{eq:corruption-process}
q_{t+1|t}(x_{t+1} \mid x_t)
= \mathcal{N}\big(x_t, (\sigma_{t+1}^2 - \sigma_t^2) \Id \big),
\end{equation}
for $t \in \intset{0}{T-1}$. This construction yields the closed-form marginal
$q_{t|0}(x_t \mid x_0) = \mathcal{N}(x_t; x_0, \sigma_t^2 \Id)$,
or equivalently $x_t = x_0 + \sigma_t \varepsilon$ with $\varepsilon \sim \mathcal{N}(0,\Id)$. The resulting joint distribution over the diffusion path factorizes as
\begin{equation*}
\textstyle
p_{0:T}(x_{0:T})
= p_0(x_0)\prod_{t=0}^{T-1} q_{t+1|t}(x_{t+1}\mid x_t).
\end{equation*}
The same joint distribution also admits a backward factorization \citep{cappe2005inference}
\begin{equation*}
\textstyle
p_{0:T}(x_{0:T})
= \Big[\prod_{t=0}^{T-1} p_{t|t+1}(x_t \mid x_{t+1})\Big]\, p_T(x_T),
\end{equation*}
where, for sufficiently large $\sigma_T$, the terminal distribution $p_T$ is well approximated by $\mathcal{N}(0,\sigma_T^2 \Id)$. The reverse transition kernels $p_{t|t+1}$ are generally intractable, as they depend on the denoising posterior
$p_{0|t}(x_0 \mid x_t) \propto p_0(x_0)\, q_{t|0}(x_t \mid x_0)$.
Indeed, $p_{t|t+1}$ can be expressed using the latter as
$$
p_{t|t+1}(x_t|x_{t+1})\!= \mathbb{E}_{x_0 \sim p_{0|t}(\cdot|x_{t+1})}\big[  q_{t|0,t+1}(x_t|x_0,x_{t+1}) \big],
$$
where the conditional transition $q_{\ell|0,t}$ for $0<\ell<t$ admits a closed-form expression \citep{ho2020ddpm,song2021ddim}. In the VE setting, it is Gaussian with mean given by a convex combination of $(x_0,x_t)$,
\begin{equation*}
q_{\ell|0,t}(x_\ell \mid x_0, x_t)
\!=\!
\mathcal{N}\!\Big(
\gamma_{\ell|t} x_t + (1-\gamma_{\ell|t}) x_0,\;
\sigma_\ell^2 (1-\gamma_{\ell|t}) \Id \!
\Big),
\end{equation*}
where $\gamma_{\ell|t} = \sigma_\ell^2 / \sigma_t^2$.
The commonly adopted surrogate for the intractable reverse kernel $p_{t|t+1}$ replaces the unknown clean sample $x_0$ in the above transition with the output of a neural network $(t, x_t)\mapsto D^\theta_{t}(x_t)$ parameterized by $\theta$. This yields the approximate reverse process
\begin{equation*}
p^\theta_{t|t+1}(x_t \mid x_{t+1})
=
q_{t|0,t+1}\big(x_t \mid D^\theta_{t+1}(x_{t+1}), x_{t+1}\big).
\end{equation*}
The neural network is trained to predict the posterior mean $D_t(x_t)\coloneqq \int x_0 \, p_{0|t}(x_0|x_t) \rmd x_0$ by minimizing a denoising regression objectives \citep{ho2020ddpm,nichol2021improved,karras2022edm}.
After training, new samples are generated by initializing $x_T \sim \mathcal{N}(0,\sigma_T^2 \Id)$ and iteratively drawing samples from $p^\theta_{t|t+1}(\cdot|x_{t+1})$ down to $x_0$.

\subsection{Diffusion Model Prior on OpenMRG Data}

We train a DM using the EDM framework \citep{karras2022edm} on the OpenMRG dataset \citep{andersson2022openmrg}, which provides synchronized CML, radar, and RG measurements over Gothenburg during June--August 2015; refer to \Cref{sec:real-dataset} for details.
Training is performed on $21{,}196$ radar-derived rain-rate fields, preprocessed with \texttt{PyNNcml} \citep{PyNNcml2026github} and discretized on a $48 \times 36$ spatial grid.
Additional implementation details are provided in Appendix~\ref{apdx:details-dm}.


\emph{Evaluation of rain field priors.}\quad
We evaluate the quality of the learned DM prior independently of the inverse problem, focusing on its ability to generate realistic rain fields.
\Cref{fig:generated-samples} presents qualitative samples drawn from the DM alongside reference rain fields, while \Cref{fig:stats-dm-prior} reports histograms of rain rates and cumulative precipitation.
Both statistics show close agreement with the reference data, and the qualitative examples indicate that the trained DM preserves the sparse structure characteristic of rainfall fields.

To further assess distributional fidelity, we use a classifier two-sample test \citep{lopez2016c2t}. Specifically, it consists of training a classifier to distinguish between a held-out set of real rain fields from generated samples by the diffusion prior. Classification accuracy close to $50\%$ indicates that generated samples are difficult to distinguish from real data, and therefore that the learned prior captures the target distribution well. We use the discriminator architecture of \citet{habi2021raingan} and report classification accuracy on a balanced held-out test set as the evaluation metric. The trained discriminator achieves $57\%$ accuracy, where chance level is $50\%$, indicating only limited distinguishability between generated and real rainfall fields. We provide more implementation details in \Cref{apdx:discriminator-details}.

\begin{figure*}[!t]
    \centering 
    \includegraphics[width=\textwidth]{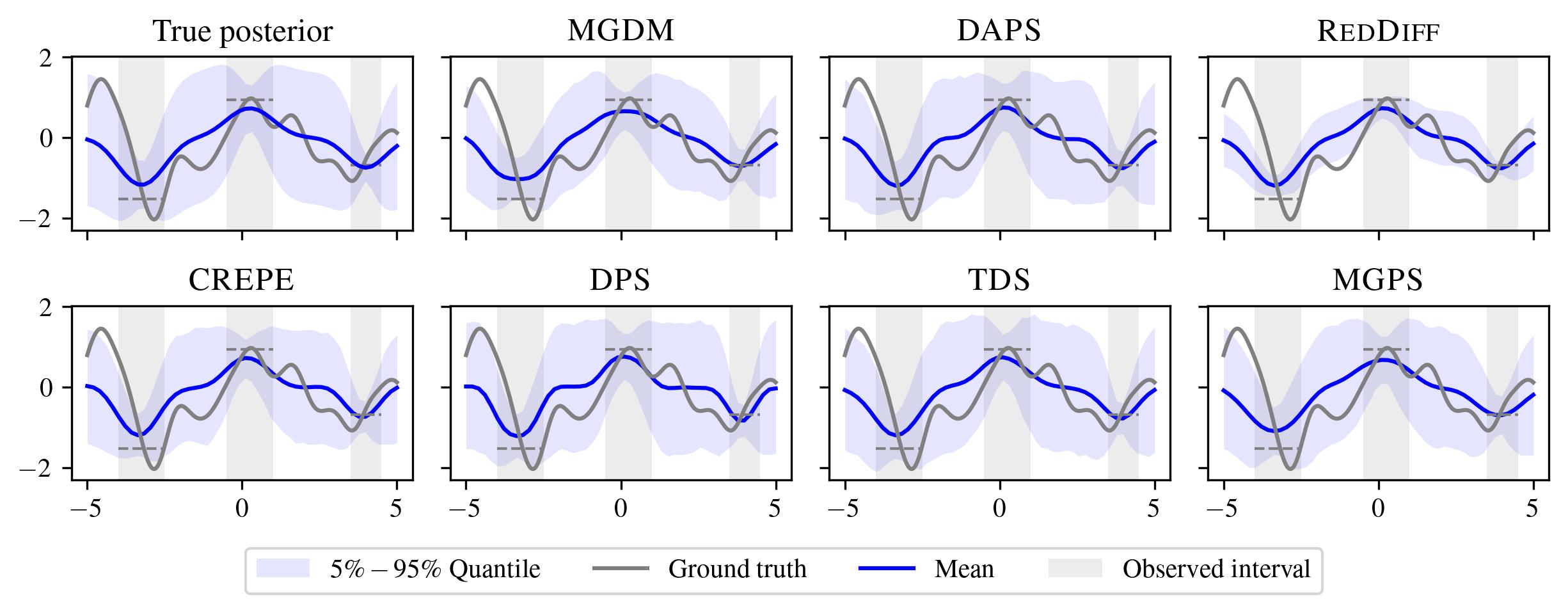}
    \captionsetup{font=small}
    \caption{Comparison between the reconstructions of the baselines on the inverse problem with diffusion prior on the setting of GP.
    The y-axis shows intensities while the x-axis represents a one-dimensional grid of the $[-5, 5]$ interval.
    The dashed horizontal lines depict values of the integral over the observed intervals.}
    \label{fig:gp-exp-viz-1}
\end{figure*}

\section{Posterior Sampling with Diffusion Models}
We now formulate CML-based rainfall reconstruction as a Bayesian inverse problem and describe training-free posterior sampling methods that leverage pre-trained DM prior.
\paragraph{CML Measurement Operator.}

Let $X:\Omega\to\mathbb{R}_+$ be a rain field and let $x_0\in\mathbb{R}_+^{H\times W}$ denote its discretization on grid points $\{\mathbf{s}_k\}_{k=1}^{HW}$, with entries $[x_0]_k = X(\mathbf{s}_k)$. 
The measurement of the CML at the $i$-link $\mathcal{L}_i$ is described by a power-law parameters $(a,b)$, and a noise level $\sigma_i$ as
\begin{equation}
    Y_i = a \int_{\mathbf{s}\in\mathcal{L}_i} X(\mathbf{s})^b \,\rmd \mathcal{L}_i \;+\; \sigma_i Z_i,
    \quad Z_i\sim\mathcal N(0,1).
\end{equation}
Under the standard piecewise-constant assumption on grid cells \citep{eshel2021quantitative}, the line integral can be approximated by a weighted sum
\begin{equation}
\label{eq:weighted-sum-siddon}
    Y_i
    = a \sum_{k=1}^{HW} \Delta^{i}_k \, 
    [x_0]_k^b
    + \sigma_i Z_i
    \ = \ \langle a\Delta^{i}, x_0^b\rangle + \sigma_i Z_i,
\end{equation}
where \emph{abuse the notation} and use $x_0^b$ to denote the elementwise $b$-th power, $\Delta^{i}_k$ is the intersection length between the link path $\mathcal L_i$ and cell $k$, and
$\langle\cdot,\cdot\rangle$ is the Frobenius inner product.
Stacking the $m$ links gives the observation model $y=\mathcal M(x_0)+\Sigma z$ with $\Sigma=\mathrm{diag}(\sigma_1,\dots,\sigma_m)$, $z\sim\mathcal N(0, I_m)$ and
$\mathcal M(x_0)_i=\langle a\Delta^{i},x_0^b\rangle$.
We compute $\{\Delta^{i}\}_{i=1}^m$ by adapting Siddon's ray-tracing algorithm \citep{siddon1985fast-tomo}; see Appendix~\ref{apdx:siddon-algo} for details.
Therefore, the likelihood of the inverse problem is
\begin{align}
\label{eq:llh-inverse-problem}
p(y\mid x_0)\propto \exp\{-\tfrac12\|y-\mathcal M(x_0)\|_{\Sigma^{-2}}^2\},
\end{align}
with $\|v\|_{\Sigma^{-2}}^2:=v^\top\Sigma^{-2}v$.
While the operator $\mathcal{M}$ is nonlinear in $x_0$, it is linear in the transformed variable $u=x_0^b$.

\paragraph{Training-Free Posterior Sampling with DM Priors.}
\label{sec:training-free-ps}
In training-free posterior sampling, the goal is, given measurements $y$, sample from the posterior distribution
\begin{equation}
    \label{eq:target-posterior}
    \pi_0(x_0 \mid y) \propto p(y \mid x_0)\, p_0(x_0),
\end{equation}
where $p_0$ is the implicit prior induced by a pre-trained DM.
Let $\{\pi_t\}_{t=0}^T$ be the marginals obtained by applying the same corruption process
$q_{t\mid 0}(\cdot \mid x_0)$, as defined in \eqref{eq:corruption-process}, to $\pi_0(\cdot \mid y)$.
Then the posterior reverse transitions satisfy
\begin{gather}
    \label{eq:posterior-transition}
    \pi_{t\mid t+1}(x_t \mid x_{t+1}, y)
    \propto
    p_t(y \mid x_t)\, p_{t\mid t+1}(x_t \mid x_{t+1}),
    \\
    \text{where}\quad
    p_t(y \mid x_t)
    =
    \mathbb{E}_{x_0 \sim p_{0\mid t}(\cdot \mid x_t)}
    \big[\, p(y \mid x_0) \,\big],
    \nonumber
\end{gather}
see \citet[Eq.~1.18]{janati2025dm-survey}.
The intermediate likelihoods $p_t(y \mid x_t)$ are typically intractable, as they involve an expectation of the likelihood under the denoising posterior distribution $p_{0\mid t}(x_0 \mid x_t)$, which is costly to sample.

A common approximation, introduced and used in the seminal work of \citet{chung2023dps}
replaces the full posterior expectation in \eqref{eq:posterior-transition} with a point estimate: one plugs the denoiser $D_t(x_t)$ into the likelihood; \emph{i.e.}
$$
p_t(y \mid x_t) \approx p_0(y \mid D_t(x_t)),
$$ 
and uses the resulting gradient to guide sampling, analogous to classifier guidance~\citep{dhariwal2021diffusion}.
This approximation is exact when $p_{0\mid t}(\cdot\mid x_t)$ concentrates at a single point which holds approximately at low noise levels but deteriorates as $\sigma_t$ increases.

Alternatively, several approaches have been proposed to alleviate the problem of intractability using different approximations.
\emph{Score-based methods}~\citep{song2022solving} modify the reverse SDE by adding a user-chosen
likelihood score term.
\emph{Projection methods}, such as DDRM~\citep{kawar2022denoising}, exploit the special structure in
linear Gaussian inverse problems.
\emph{Variational approaches}~\citep{mardani2024reddiff, moufad2025mgps, janati2025mixture} use different approximation of $p_t(y \mid x_t)$ together with variational inference to approximate either the posterior $\pi_0$ in \eqref{eq:target-posterior}
(e.g., with a Gaussian)~\citep{mardani2024reddiff}, or the posterior transitions ~\citep{moufad2025mgps, janati2025mixture}.
\emph{Sequential Monte Carlo (SMC)} approximates the sequence of posterior distributions $\{\pi_t(\cdot \mid y)\}_{t=0}^T$ using a population of weighted particles \citep{delmoral2006smc,chopin2020introduction}.
Unlike variational approaches that evolve a single solution, SMC propagates and reweights $N$ particles, thereby maintaining a population-based representation of the posterior.
\emph{Replica Exchange} constructs a Markov chain targeting the product distribution $\bigotimes_{t=0}^T \pi_t(\cdot \mid y)$.
Each iteration alternates between two types of updates:
(i) \emph{local moves}, which independently target the marginals $\pi_t(\cdot \mid y)$ and can be executed in parallel, and
(ii) \emph{communication moves}, which propose Metropolis--Hastings swaps between neighboring states $(t_{m-1}, t_m)$.
In \citet{he2025crepe}, the communication step departs from standard swap proposals by instead proposing new states.
Since the resulting acceptance ratio involves the prior marginal density, \citet{he2025crepe} relies on the estimator of \citet{he2025rne}, which computes density ratios via the forward and the surrogate reverse processes.

We refer to \citet{daras2024dm-survey, janati2025dm-survey} for detailed survey on the landscape of methods for training-free posterior sampling with DM priors.

\begin{figure*}[t]
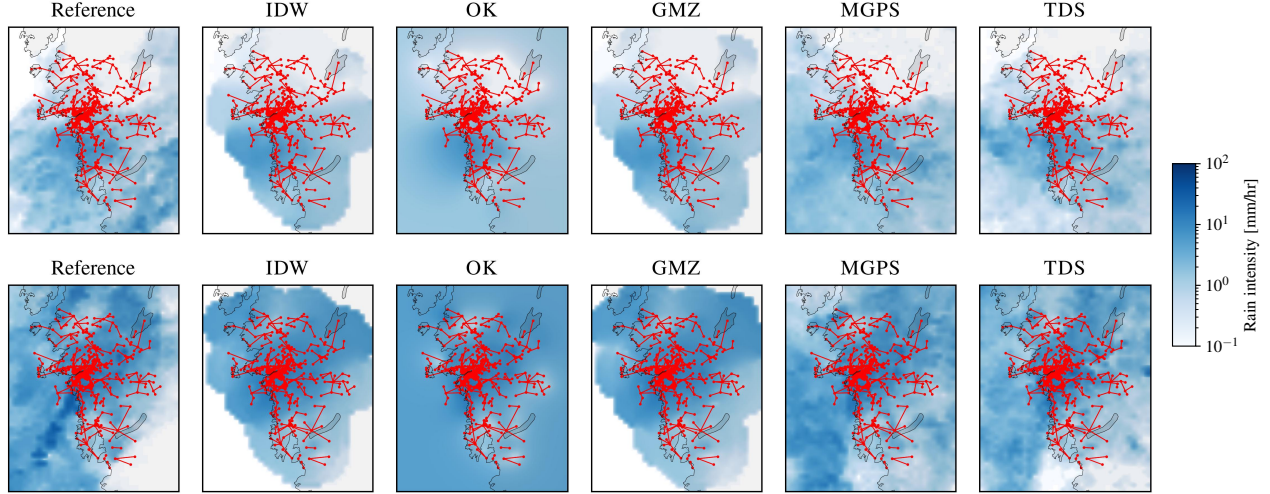

    \centering
    \foreach \method in {ref,idw,ok,gmz,mgps,tds}{%
        \begin{subfigure}[t]{0.15\textwidth}
            \centering
            \includegraphics[width=\textwidth]{files/figs/rain_recs_real_jpeg/3403/\method.jpeg}
        \end{subfigure}%
    }%
    \begin{subfigure}[t]{0.055\textwidth}
        \centering
        \hspace{0.055\textwidth}
    \end{subfigure}%
    
    \centering
    \foreach \method in {ref,idw,ok,gmz,mgps,tds}{%
        \begin{subfigure}[t]{0.15\textwidth}
            \centering
            \includegraphics[width=\textwidth]{files/figs/rain_recs_real_jpeg/3370/\method.jpeg}
        \end{subfigure}%
    }%
    \begin{subfigure}[t]{0.055\textwidth}
        \centering
        \vspace{-47mm} 
        \includegraphics[width=1.5\textwidth]{files/figs/rain_color_bar.pdf}
    \end{subfigure}%
    \captionsetup{font=small}
    \caption{Comparisons of rain field reconstructions on real CMLs links from OpenMRG. We depict the network of CMLs in red.}
    \label{fig:total_maps}
\end{figure*}

\section{Experiments}
\paragraph{Experimental Settings.}
We conduct (i) simulated and controlled experiments based on GP prior \citep{rasmussen2006gp}, and (ii) real-world experiments on the OpenMRG dataset \citep{andersson2022openmrg,fencl2024data}.
In our target setting, CMLs are the primary source of observations.
Hence, we consider as classical baselines \idw, \gmz, and \ok, whereas \RM\ is not applicable because it requires rain-gauge measurements to estimate the marginal distributions used by the method.

\paragraph{Meteorological Baselines.} We briefly revisit the principles behind the baselines used in rain field reconstruction.

\emph{Inverse Distance Weighting (\idw).}\quad
In IDW, the path-integrated observations of CMLs are converted into VRGs at the middle of the links. The rain rate at location $\mathbf{s}$ is estimated as a weighted average of nearby observations
\begin{equation*}
\textstyle
\hat{X}(\mathbf{s})=\sum_{i=1}^{N} w_i(\mathbf{s})\,X(\mathbf{s}_i),
\quad
w_i(\mathbf{s})= d_i^{-p} / \sum_{j=1}^{N} d_j^{-p},
\end{equation*}
where $X(\mathbf{s}_i)$ is the rainfall intensity associated with the $i$-th VRG positioned at location $\mathbf{s}_i$ , $d_i=\|\mathbf{s} - \mathbf{s}_i\|$ is the distance to $\mathbf{s}_i$, and $p>0$ controls the decay of influence with distance.

\emph{\gmz\ algorithm.}\quad
Rather than representing each CML by a single VRGs at its midpoint, the GMZ algorithm accounts for the variability of the rainfall rate along the CML link considering $K$ VRGs evenly distributed along the CML link.
More precisely, for the CML link $\mathcal{L}_i$, the $K$ VRGs are positioned at $\{\mathbf{s}_{i,k}\}_{k=1}^K$, and the rain intensity at the latter points is constrained to match the observation $Y_i = \frac{a}{K}\sum_{k=1}^{K} X_{\ell,k}^b$. This redistribution is performed iteratively using neighboring links to enforce spatial consistency, and the resulting VRGs are interpolated with IDW \citep{goldshtein2009idw-gmz}.

\emph{Ordinary Kriging (\ok).}\quad
Similar to IDW, OK uses the VRG simplification of the path-integrated measurement.
OK uses GP prior with user-defined kernel (variogram) as an assumption over the rain field to interpolate the observations.
The prior parameters are chosen by minimizing the L1 norm \citep{matheron1963geostatistics,cressie1993spatialdata}.

\paragraph{Diffusion Baselines.}
We select seven training-free posterior sampling baselines that represent the approach presented in \Cref{sec:training-free-ps}, namely,
\dps~\citep{chung2023dps},
\daps~\citep{zhang2024daps},
\reddiff~\citep{mardani2024reddiff},
\crepe~\citep{he2025crepe},
\mgps~\citep{moufad2025mgps},
\mgdm~\citep{janati2025mixture},
\tds~\citep{wu2023tds}.

We provide details on the implementation of both meteorological and Diffusion baselines as well as hyperparameter choices in \Cref{apdx:implem-details}.
When reporting results, we highlight the best metric with \first{\transparent{0}{tx}} and second best \second{\transparent{0}{tx}}.

While \crepe\ is included in the Gaussian process experiments, we exclude it from the CML experiments.
In this setting, the method exhibits substantially higher computational cost than the other baselines and yields moderate performance.
For these reasons, we omit \crepe\ from both the quantitative results and qualitative comparisons.

\subsection{Simulated Examples}
We consider a one-dimensional linear inverse problem where the latent field is a Gaussian process
$X \sim \mathcal{GP}(0, k)$ on $[-5, 5]$, with an RBF kernel
\begin{equation}
k(s, s') = \exp\!\left(-\frac{(s - s')^2}{2\ell^2}\right),
\qquad \ell = 0.6.
\end{equation}
The links $\{\mathcal{L}_i\}_{i=1}^m$ in this setting reduce to intervals $\{[a_i, b_i]\}_{i=1}^m$\footnote{In the GP experiments, we abuse the notation and use $a_i$ and $b_i$ to denote integration intervals instead of the power-law constants.}
,and the observation model 
\begin{equation}
Y_i=\int_{a_i}^{b_i} X(s)\, \mathrm{d}s + \sigma Z_i,
\qquad Z_i \sim \mathcal{N}(0, 1),
\end{equation}
where $\{Z_i\}_{i=1}^m$ are mutually independent and independent of $X$.
Because the observation operator is linear and the prior is Gaussian, the posterior is a GP with closed-form mean and covariance \citep{rasmussen2006gp}:
\begin{align*}
\mu_{X|y}(s) &= \mathbf{k}_y(s)^\top (\mathbf{K}_{yy} + \sigma^2 \mathbf{I})^{-1} \mathbf{y}, \\
k_{X|y}(s, s') &= k(s, s') - \mathbf{k}_y(s)^\top (\mathbf{K}_{yy} + \sigma^2 \mathbf{I})^{-1} \mathbf{k}_y(s'),
\end{align*}
where $\mathbf{k}_y(s)$ is a vector in $\mathbb{R}^m$ with coordinates $[\mathbf{k}_y(s)]_i = \int_{a_i}^{b_i} k(s, s')\, \mathrm{d}s'$, and $\mathbf{K}_{yy}$ is a matrix in $\mathbb{R}^{m \times m}$ with elements
$[\mathbf{K}_{yy}]_{ij} = \int_{a_i}^{b_i} \int_{a_j}^{b_j} k(s, s')\, \mathrm{d}s\, \mathrm{d}s'$.
In our case, we postpone the derivations of these formulas to \Cref{apdx:gp-exp}.
We discretize the interval $[-5, 5]$ into $50$ points.
Evaluating the Gaussian process $X$ on this grid yields a multivariate Gaussian distribution.
Consequently, for training-free baselines, we employed a DM prior to targeting this Gaussian distribution.
Such a DM admits a closed-form construction, as described in \citet[Appendix~B]{moufad2025mgps}.

\paragraph{Evaluation.}
This setting provides an oracle posterior, enabling a precise evaluation.
We report (i) the sliced Wasserstein distance between the true and approximate posteriors \citep{rabin2011sliced, bonneel2015sliced}, and
(ii) $\ell_2$ errors between the oracle and the approximate quantities: posterior mean and the $5\%$-$95\%$ pointwise quantiles.

\paragraph{Results.}
The results are reported in \Cref{table:gp-metrics}.
On the GP benchmark, TDS is the strongest overall method, achieving the best distributional match (lowest SW) and the most accurate uncertainty while remaining competitive on the mean.
In contrast, \reddiff\ achieves the lowest mean error but performs poorly on SW and tails, indicating that mean accuracy alone can mask severe posterior miscalibration.
\daps\ and \mgps\ perform competitively with moderate quantile errors.
\mgdm\ is generally less competitive, with a higher SW and substantially worse tail errors.
We provide qualitative examples in \Cref{fig:gp-exp-viz-1,fig:gp-exp-viz-2}.


\begin{table}[t]
    \centering
    \captionsetup{font=small}
    \caption{Quantitative comparison between Diffusion Models baselines on GP experiment. First row reports Sliced Wasserstein distance.
    The last 3 rows report $\ell_2$ distance between reference and approximate mentioned quantity. Lower metrics are better.
    }
    \resizebox{0.48\textwidth}{!}{
    \begin{tabular}{l cccc ccc}
    \toprule
    & {\mgdm} & {\daps} & {\reddiff} & {\crepe} & {\dps} & {\tds} & {\mgps} \\
    \midrule
    SW \; $\downarrow$
        & 0.16 & \second{0.09} & 0.38 & 0.13 & 0.12 & \first{0.07} & \second{0.09} \\
    Mean \; $\downarrow$
        & 0.61 & 0.54 & \first{0.36} & 0.68 & 0.98 & \second{0.53} & \second{0.53} \\
    $5\%$-Quantile \; $\downarrow$
        & 2.21 & 1.12 & 5.70 & 1.82 & 1.17 & \first{0.85} & \second{0.99} \\
    $95\%$-Quantile \; $\downarrow$
        & 2.20 & \second{1.29} & 5.73 & 1.51 & 1.30 & \first{1.02} & 1.35 \\
    \bottomrule
    \end{tabular}
    }
    \label{table:gp-metrics}
\end{table}

\subsection{Real Dataset}
\label{sec:real-dataset}

We evaluate our approach on the OpenMRG dataset \cite{andersson2022openmrg} and use the Python pacakge \texttt{PyNNcml} \citep{PyNNcml2026github,chwala2026open} to download and preprocess it.
The dataset provides synchronized measurements from CMLs, weather radar, and RGs over the Gothenburg metropolitan area (Sweden) during June--August 2015.

\paragraph{Commercial Microwave Links.}
The network contains 364 bidirectional links (728 sub-links) operating at frequencies between 7 and 38 GHz (mostly above 25 GHz), with path lengths between 0.1\, km and 15\, km (median $\approx 2$\, km). 
Transmitted and received signal levels are recorded every 10 seconds for each sub-link. 
Each link is associated with a pair of power-law parameters $(a, b)$ (Eq.~\ref{eq:power-law}), whose dependence on frequency and path length is summarized in \Cref{fig:cmls-power-law-stats}.


\paragraph{Radar-derived Reference Fields.}
Radar data are drawn from the Swedish NORDRAD composite product operated by SMHI. 
The composite is generated from multiple Swedish radars using the Rainbow software and provides reflectivity fields on a 2 km grid; coverage over Gothenburg is primarily from the Vara radar (at $\approx$ 78 km), with partial contributions from nearby radars. 
The product is gauge-corrected using a distance-dependent gauge-radar ratio estimation \cite{michelson2000gaugeradar}. 
Reflectivity is expressed in dBZ and stored as pseudo-dBZ values, which are converted back to actual dBZ before processing; values are then mapped to linear units and converted to rainfall intensity via a standard $Z$-$R$ power-law relationship commonly used for stratiform conditions.

\paragraph{Evaluation.}
Since spatially dense ground truth is not directly observed, we use the gauge-adjusted radar-derived rain-rate fields as reference maps for evaluation.
Attenuation observations are generated by applying the CML observation operator to radar rain-rate fields from the test split, using the OpenMRG network topology, link lengths, and link-specific power-law parameters $(a_i,b_i)$.
We consider two noise settings:
\circled{1} \emph{Isotropic noise} where all CML observations are corrupted with the same noise level $\sigma$ and \circled{2} \emph{Heteroscedastic noise}, where the noise level depends on the CML length. Specifically, for link $i$ with length $L_i$, we set
$
\sigma_i = \frac{\sigma}{2}\left(1 + \frac{L_i}{L_{\max}}\right),
$
so that longer links have larger noise levels, with $\sigma_i=\sigma$ when $L_i=L_{\max}$ and $\sigma/2 \leq \sigma_i < \sigma$ otherwise.
Note that other heteroscedastic noise models could also be considered. Such settings can be incorporated straightforwardly by specifying the covariance matrix $\Sigma$ of the observation operator in \Cref{eq:llh-inverse-problem}.

\paragraph{Metrics.}
Existing studies commonly rely on standard pixel-wise reconstruction metrics \citep{graf2021rainfall,zhang2023reconstructing}. In our assessment, we report the root mean square error (RMSE), which quantifies the discrepancy between ground-truth and reconstructed fields, and the Pearson correlation coefficient (PCC), which measures their linear agreement. We also evaluate the error in cumulative rainfall, defined as the difference between the total precipitation of the reconstructed and ground-truth fields. Values closer to zero indicate better agreement, with positive values corresponding to overestimation and negative values to underestimation of total rainfall.
For computational reasons, we restrict the maximum runtime of the diffusion baselines to $3$ minutes.
For \tds, this requires reducing the number of particles to satisfy the runtime constraint. Hyperparameter details for all baselines are provided in \Cref{table:hyperparameters_ps}.

\paragraph{Results.}
\Cref{table:real-cml-openmrg-metrics} shows that the diffusion-model baselines consistently outperform the meteorological baselines across both settings and evaluation metrics.
In particular, \mgps\ achieves strong performance on all three metrics. Overall, DM-based reconstructions exhibit higher agreement with the reference fields, as reflected by their \pcc\ values. Among the meteorological methods, \gmz\ performs well in terms of cumulative rainfall, while \ok\ yields the strongest overall performance within this class of baselines. The results under heterogeneous noise are slightly better, as expected, since this setting has a lower overall noise level.

Qualitative comparisons are provided in \Cref{fig:total_maps}, with a detailed comparison including all methods in \Cref{fig:real-cmls-exp-all-comp-1,fig:real-cmls-exp-all-comp-2}.

\paragraph{Runtime.}
To assess computational cost, we report the runtimes of the meteorological and diffusion-based baselines in \Cref{table:runtime_baselines}. The results highlight a trade-off: meteorological methods are fast but they (expect \ok) produce a deterministic reconstruction; the diffusion baselines are more expensive yet they generate ensembles of plausible rain fields. This enables the assessment of multiple scenarios, including best- and worst-case precipitation patterns that may be relevant for flood-warning applications.

\begin{table}[ht]
    \centering
    \caption{Runtime (in seconds) per reconstruction for all baselines. For diffusion-based methods, the reported runtime is divided by $10$, since a batch of $10$ samples is generated jointly.}
    \resizebox{0.48\textwidth}{!}{
    \begin{tabular}{l cccccc | ccc}
    \toprule
    & \daps & \dps & \mgdm & \mgps & \reddiff & \tds & \idw & \gmz & \ok \\
    \cmidrule(lr){2-10}
    Runtime & 6.06 & 5.27 & 6.87 & 5.22 & 5.95 & 18.49 & 0.01 & 1.01 & 0.27 \\
    \bottomrule
    \end{tabular}
    }
    \label{table:runtime_baselines}
\end{table}

In our experiments, generating $10$ samples with the diffusion baselines takes roughly one minute. Since CML data is aggregate at intervals of $5$ to $15$ minutes, these runtimes remain compatible with real-time meteorological monitoring.

\begin{table}[t]
    \centering
    \captionsetup{font=small}
    \caption{Comparison between DM and meteorological baselines on \rmse, \pcc, and difference of Cumulative Rain on real CML links from OpenMRG. 95\%-confidence intervals are shown.}
    \resizebox{0.47\textwidth}{!}{
    \begin{tabular}{l ccc}
    \toprule
    & \rmse\,$\downarrow$ & \pcc\,$\uparrow$ &  Cumulative Rain $\rightarrow0$  \\
    \midrule
    & \multicolumn{3}{c}{Isotropic noise} \\
    \cmidrule(lr){2-4}
    \daps & 0.88 $\pm$ \small{0.09} & \second{0.24} $\pm$ \small{0.02}& -320 $\pm$ \small{53.66} \\
    \dps & 0.88 $\pm$ \small{0.09} & 0.20 $\pm$ \small{0.01} & -333 $\pm$ \small{54.34} \\
    \mgdm & 0.91 $\pm$ \small{0.08} & 0.22 $\pm$ \small{0.02} & 231 $\pm$ \small{35.29} \\
    \mgps & \second{0.86} $\pm$ \small{0.08} & \first{0.24} $\pm$ \small{0.01} & \first{42} $\pm$ \small{34.65} \\
    \reddiff & 0.86 $\pm$ \small{0.09} & 0.23 $\pm$ \small{0.02} & -356 $\pm$ \small{48.47} \\
    \tds & \first{0.84} $\pm$ \small{0.09} & 0.21 $\pm$ \small{0.02} & -216 $\pm$ \small{35.89} \\
    \cmidrule(lr){2-4}
    \idw & 0.90 $\pm$ \small{0.09} & 0.21 $\pm$ \small{0.01} & -144 $\pm$ \small{32.26} \\
    \gmz & 0.92 $\pm$ \small{0.09} & 0.21 $\pm$ \small{0.01} & \second{-65} $\pm$ \small{29.26} \\
    \ok & 0.87 $\pm$ \small{0.09} & 0.23 $\pm$ \small{0.02} & 158 $\pm$ \small{38.67} \\
    \cmidrule(lr){2-4}
    & \multicolumn{3}{c}{Heteroscedastic noise} \\
    \cmidrule(lr){2-4}
    \daps & 0.87 $\pm$ \small{0.10} & 0.24 $\pm$ \small{0.01} & -438.80 $\pm$ \small{76.08} \\
    \dps & 0.85 $\pm$ \small{0.09} & 0.21 $\pm$ \small{0.01} & -349.25 $\pm$ \small{53.49} \\
    \mgdm & 0.92 $\pm$ \small{0.08} & 0.24 $\pm$ \small{0.01} & 300.00 $\pm$ \small{36.24} \\
    \mgps & \second{0.83} $\pm$ \small{0.08} & \first{0.26} $\pm$ \small{0.01} & \first{70.94} $\pm$ \small{36.40} \\
    \reddiff & 0.84 $\pm$ \small{0.09} & 0.24 $\pm$ \small{0.02} & -370.92 $\pm$ \small{48.78} \\
    \tds & \first{0.81} $\pm$ \small{0.09} & \second{0.25} $\pm$ \small{0.02} & -241.43 $\pm$ \small{35.03} \\
    \cmidrule(lr){2-4}
    \idw & 0.86 $\pm$ \small{0.09} & 0.24 $\pm$ \small{0.01} & -166.95 $\pm$ \small{32.09} \\
    \gmz & 0.91 $\pm$ \small{0.09} & 0.21 $\pm$ \small{0.01} & \second{-81.93} $\pm$ \small{29.09} \\
    \ok & 0.86 $\pm$ \small{0.09} & 0.24 $\pm$ \small{0.02} & 140.74 $\pm$ \small{38.83} \\
    \bottomrule
    \end{tabular}
    }
    \label{table:real-cml-openmrg-metrics}
\end{table}

\subsection{Ablation}

\paragraph{DM Prior and Nonlinear Observation Model.}
We first disentangle the gains due to the diffusion prior from those due to accurate modeling of the nonlinear path-integrated observation operator. To this end, we consider an ablation in which observations are generated under the true nonlinear setting, with power law $b_i>1$, while inference assumes $b_i=1$, yielding a linear observation model. To assess the role of the diffusion prior quality, we repeat this experiment with both a \emph{``Strong''} and a \emph{``Weak''} prior; the latter corresponding to an undertrained diffusion model trained for a small number of epochs. Results are reported in \Cref{table:ablation-dm-and-operator}.

\emph{Discussion.}\quad
The results show that the diffusion prior improves the reconstructed rain fields; with accurate modeling of the nonlinear observation operator being essential for strong quantitative performance.

\begin{table}[h]
    \centering
    \captionsetup{font=small}
    \caption{Ablation between DMs and meteorological baselines on \rmse, \pcc, and difference of Cumulative Rain (Cum Rain).
    Two settings are considered on OpenMRG dataset with simulated CMLs: few-long and many-short links, with the power-law $b=1$. For \rmse\ lower is better. For \pcc, higher is better. For Cum Rain, closer to 0 is better.}
    \resizebox{0.48\textwidth}{!}{
    \begin{tabular}{l ccc | ccc}
    \midrule
    & \multicolumn{3}{c}{Few-long links}      & \multicolumn{3}{c}{Many-short links} \\
    \cmidrule(lr){2-7}
    & \rmse\,$\downarrow$ & \pcc\,$\uparrow$ &  Cum Rain  & \rmse\,$\downarrow$ & \pcc\,$\uparrow$ &  Cum Rain   \\
    \midrule
    \daps & 0.92 & 0.24 & -243      & 1.16 & 0.29 & -348 \\   
    \dps & 0.88 & 0.23 & -267       & 0.80 & 0.32 & -152 \\  
    \mgdm & 0.89 & 0.25 & 152       & 0.79 & 0.38 & 153 \\  
    \mgps & \second{0.84} & \second{0.28} & \first{11}        & \first{0.75} & 0.41 & 20 \\ 
    \reddiff & 0.85 & 0.26 & -221   & 0.79 & 0.37 & -224 \\      
    \tds & 0.85 & 0.23 & -163       & 0.76 & 0.34 & -77 \\  
    \cmidrule(lr){2-7}
    \idw & 0.89 & 0.25 & -164       & \first{0.75} & \second{0.42} & \first{3} \\  
    \gmz & 0.86 & \first{0.29} & -20        & \first{0.75} & \first{0.43} & \second{15} \\ 
    \ok & \first{0.83} & \second{0.28} & \second{19}          & 0.76 & 0.38 & 27 \\ 
    \bottomrule
    %
    \end{tabular}
    }
    \label{table:ablation}
\end{table}

\paragraph{CML Network Configuration.}
We next assess the impact of CML network geometry, focusing on the number and spatial configuration of links. We consider two synthetic configurations: a \emph{few-long} setting with $25$ long CMLs and a \emph{many-short} setting with $100$ short CMLs. To isolate the effect of spatial configuration, we restrict this ablation to the case where the power-law exponent is equal to $1$, resulting in a linear inverse problem. Results are reported in \Cref{table:ablation}.

\emph{Discussion.}\quad
In this idealized linear setting, meteorological baselines perform comparably to DM-based methods, with \ok\ achieving the lowest \rmse. This suggests that, under linear sensing and sufficiently dense coverage, the rainfall field can be well approximated by a Gaussian-process model and effectively reconstructed by interpolation. Meteorological baselines further benefit from increased CML density, particularly when links are well distributed across the domain, as shown in \Cref{table:ablation}.

Real-world CML observations, however, deviate substantially from this regime. As shown in \Cref{fig:cmls-power-law-stats}, the power-law exponent $b$ varies over a wide range, inducing nonlinear measurements. Moreover, operational CML networks are typically spatially heterogeneous, with dense coverage in some regions and sparse or absent coverage in others, as illustrated in \Cref{fig:total_maps}. These factors highlight the limitations of interpolation-based meteorological baselines and motivate the use of expressive priors coupled with physically grounded observation models, as in the proposed DM-based framework.

\begin{table}[h]
    \centering
    \captionsetup{font=small}
    \caption{Ablation to discern the improvement that comes from the DM prior vs. the improvement from accounting for the nonlinearity of the measurement operator. \emph{``Weak-prior''} is an under-trained DM, i.e. a DM trained with very-few number epochs.
    }
    \resizebox{0.48\textwidth}{!}{
    \begin{tabular}{ll ccc | ccc}
    \toprule
    & & \multicolumn{3}{c}{Linear operator} & \multicolumn{3}{c}{Nonlinear operator} \\
    \cmidrule(lr){3-5} \cmidrule(lr){6-8}
    DM & Sampler & \rmse\,$\downarrow$ & \pcc\,$\uparrow$ &  Cum Rain $\rightarrow0$  &\, \rmse\,$\downarrow$ & \pcc\,$\uparrow$ &  Cum Rain $\rightarrow0$ \\
    \midrule
    \multirow{3}{*}{Weak} 
    & \multicolumn{1}{l|}{\mgps}    & 0.94 & 0.17 & -430.30 &\, 0.88 & 0.23 & 132.21 \\
    & \multicolumn{1}{l|}{\reddiff} & 1.01 & 0.16 & -661.38 &\, 0.90 & 0.19 & -456.52 \\
    & \multicolumn{1}{l|}{\tds}     & 0.96 & 0.11 & -543.51 &\, 0.86 & 0.18 & -259.54 \\
    \midrule
    \multirow{3}{*}{Strong}
    & \multicolumn{1}{l|}{\mgps}    & 0.96 & 0.16 & -436.11 &\, 0.86 & 0.24 & 42.86 \\
    & \multicolumn{1}{l|}{\reddiff} & 1.00 & 0.17 & -643.39 &\, 0.86 & 0.23 & -356.76 \\
    & \multicolumn{1}{l|}{\tds}     & 0.97 & 0.13 & -549.76 &\, 0.84 & 0.21 & -216.44 \\
    \bottomrule
    \end{tabular}
    }
    \label{table:ablation-dm-and-operator}
\end{table}

\section{Conclusion}
We presented a Bayesian framework for rain field reconstruction from CMLs that leverages DMs as expressive spatial priors. Formulating the problem as a Bayesian inverse problem enables training-free posterior sampling using a wide range of modern inference methods. Crucially, the proposed framework incorporates a physically grounded measurement operator that accounts for the path-integrated nature of CML observations, moving beyond the VRG assumption.

\paragraph{Limitations and Future Work.}
The DM prior is trained on data from a single geographic region and season. 
Its performance under distribution shift to other climates and meteorological regimes has not been assessed.
Moreover, our current formulation treats rainfall fields independently across time, thereby neglecting temporal structure.
Future work will explore spatiotemporal modeling extensions to address these limitations.




\section*{Impact Statement}


The proposed approach may contribute to the
monitoring of hydric resources by enabling more effective use of opportunistic
sensing data, and may support urban flood and hazard prevention through
improved rainfall estimation.
At the same time,
because opportunistic sensing relies on telecommunications infrastructure, its
deployment may introduce user exposure and privacy-related risks that warrant
careful consideration. These potential impacts should be taken into account
when translating such methods into operational systems.

\section*{Acknowledgements}

The work of Badr Moufad has been supported by Technology Innovation Institute (TII), project Fed2Learn. The work of Eric Moulines has been partly funded by the European Union (ERC-2022-SYG-OCEAN-101071601). Views and opinions expressed are however those
of the author(s) only and do not necessarily reflect those of the European Union or the European Research Council Executive Agency. Neither the European Union nor the granting authority can be held responsible for them. This work was granted access to the HPC resources of IDRIS under the allocations 2025-AD011015980 made by GENCI.





\bibliography{ref.bib,ref_sota.bib,ref_overleaf.bib}
\bibliographystyle{icml2026}

\newpage
\appendix
\onecolumn 
\section{Extended Related Work}
\label{sec:extended_related_work}

As discussed in Section~\ref{sec:related work}, earlier studies on rainfall reconstruction from CMLs \cite{leijnse2007rainfall, goldshtein2009idw-gmz, overeem2013country, overeem2016retrieval} typically reduced each path-integrated link observation to a VRG located at the link midpoint and then produced gridded rain fields via deterministic interpolation (most commonly \idw\ or \ok).
This abstraction enabled the first large-scale deployments of CML-based mapping, but it discards variability along the link and effectively converts line-integral measurements into point-scale information, which can yield overly smooth fields and miss localized extremes.

Recent work has systematically examined the implications of this approximation and the extent to which more faithful representations of CML measurements can improve reconstruction accuracy. 
\citet{messer2022rain} quantify the information gain of line-integral observations relative to point measurements, demonstrating that CMLs contain additional spatial information that is lost when collapsed to a single point. 
Building on this insight, \citet{eshel2021quantitative} evaluate rainfall reconstruction using a dense operational CML network in southern Germany and a semisynthetic framework based on gauge-adjusted radar data. Their results confirm that reconstruction accuracy strongly depends on rainfall spatial structure, improving for smoother fields with larger decorrelation distances. 
Importantly, they show that representing each CML by multiple VRGs using GMZ yields a consistent, though modest, improvement over the single-midpoint VRG representation, largely independent of whether \idw\ or \ok\ is applied.

Similar conclusions emerge from other deterministic reconstruction analyses. \citet{zhang2023reconstructing} investigate two-dimensional rainfall fields reconstruction by converting attenuation measurements into path-averaged rain rates via power-law relationships and then interpolating them with \idw\ or \ok. Their analysis shows that both methods recover the dominant spatial structures, with \ok\ typically outperforming \idw\ due to its explicit modeling of spatial dependence. However, reconstruction quality deteriorates in sparsely covered regions and during high-intensity events, reflecting fundamental limitations imposed by network geometry and path integration. 
Moving beyond CML-only settings, \citet{graf2021rainfall} demonstrate that jointly assimilating rain gauges, personal weather stations, and CMLs within a block-kriging framework substantially increases effective observation density and improves representation of spatial rainfall variability. Crucially, their approach accounts for differences in spatial support and sensor-specific uncertainties, yielding performance comparable to or exceeding gauge-adjusted radar at local and regional scales. 
Likewise, \citet{pasierb2024application} emphasize the sensitivity of CML-based reconstruction to link characteristics (e.g., length, frequency, orientation) as well as to the strategy used to project path-averaged measurements onto spatial grids, underscoring that preprocessing and mapping choices play a critical role in reconstruction quality.

While deterministic interpolation provides a single best estimate, it offer limited uncertainty quantification and may further oversmooth fine-scale variability when constraints are weak. To address these limitations, stochastic reconstruction approaches treat rainfall as a random spatial process conditioned on both point and path-integrated information. A key advance in this direction is the application of the RM framework by \citet{haese2017stochastic}, which generates an ensemble of rainfall fields that agree with rain gauges and with the path-averaged link measurements, so it keeps the fact that CMLs measure rainfall along a path rather than at a point. Compared to \ok, RM can better preserve small-scale spatial variability and represent higher intensities (including extreme thresholds). However, because the marginal rainfall distribution is derived primarily from gauge data, RM remains sensitive to gauge representativeness and may underestimate extremes or delay rainfall onset when gauges inadequately sample localized events.

The RM framework has since been extended to larger spatial domains. \citet{blettner2022combining} extend stochastic reconstruction to national domains by combining rain gauges and CMLs across an entire country, improving agreement with gauge-adjusted radar and reducing structural errors in rainfall pattern representation compared to deterministic kriging.
Their uncertainty decomposition (amplitude, structure, location) highlights the diagnostic value of ensemble reconstructions. Nonetheless, increased computational cost, interpretative complexity, and the smoothing implied by path-integrated constraints continue to limit the recoverable resolution of fine-scale precipitation features, particularly where observational coverage is sparse.
\section{Prior Diffusion Model}
\label{apdx:details-dm}

We base the prior model on the EDM framework \cite{karras2022edm} and adapt the publicly released implementation\footnote{\url{https://github.com/NVlabs/edm}} to our setting.

\paragraph{Dataset.}
The OpenMRG dataset contains $26{,}495$ rain-rate fields, which are split chronologically into training and test sets with a $0.8/0.2$ ratio, yielding $21{,}196$ and $5{,}299$ samples, respectively.
Unlike the original EDM setup, we operate directly on rain-rate fields without converting them to PNG images.
Each field has shape $(1,48,36)$; the width is cropped from $37$ to $36$ for consistency.
To mitigate the long-tailed distribution of rainfall intensities, we experimented with two approaches:

\emph{Log-transform.}
We initially experimented with training on a log-transformed version of the rainfall data, which amounts to training a latent diffusion model where the encoder applies a log-transform and the decoder applies an exponential transform.
While training in log-space was significantly faster (roughly $\times 1.5$), we ultimately opted against using the log-space model during inference.
The primary reason is that mapping back to rain intensities via an exponential transform introduces an additional nonlinearity into the inverse problem.
As noted in recent work on inverse problems with latent diffusion models, such nonlinearities can introduce numerical instabilities for diffusion samplers; in our case, the exp-transform led to instabilities for some of the diffusion baselines, namely for \dps\ and \daps.

\emph{Linear-transform with quantiles.}
All samples are normalized by a fixed constant corresponding to the $0.999$ quantile of the training data.
We retained this as our final preprocessing choice: this linear normalization mitigates skewed values while preserving linearity in the input space, which we found more stable for inference.

\paragraph{Architecture.}
We use DDPM++ backbone \citep{song2021dm-sde} (\texttt{SongUNet}), a U-Net encoder--decoder with skip connections and timestep embeddings.
The model uses \texttt{model\_channels}=32 and \texttt{dropout}=0.1.
The denoiser is wrapped with EDM preconditioning (\texttt{EDMPrecond}) and $\sigma_{\mathrm{data}}=0.079$ is used in both the preconditioner and the EDM loss.
A ReLU activation is applied to the denoiser output to enforce the physical constraint that the denoiser expectation $\mathbb{E}[X_0 \mid X_t]$ be non-negative. 

\emph{Enforcing non-negative values.}\quad
While the ReLU activation function is a valid heuristic to enforce non-negative values, it introduces a hard thresholding that can distort gradients near zero-intensity regions.
We also experimented with a smoother alternative, namely the Softplus activation function\footnote{\url{https://docs.pytorch.org/docs/stable/generated/torch.nn.Softplus.html}}, which yields smoother gradients around zero.
In practice, however, we found no significant difference in the performance of the diffusion prior between the two variants, indicating that our results are largely insensitive to whether non-negativity is enforced by a hard or smooth transformation.

\paragraph{Training Setup.}
Training is performed using Adam \citep{kingma2014adam} with $\beta_1=0.9$, $\beta_2=0.999$, and a constant learning rate of $10^{-4}$ for approximately $5{,}284$ epochs.
Data augmentation is limited to random axis flips, with each sample independently flipped along the $x$ and $y$ axes with probability $0.1$.
Exponential moving average updates are disabled.
The training is conducted on two NVIDIA H100 (80GB) GPUs, with a total training time of approximately $10$ hours.

\paragraph{Discriminator Training Details.}
\label{apdx:discriminator-details}
The discriminator is trained as a supervised binary classifier to distinguish real from generated samples, with labels ``1'' and ``0'', respectively.
The two datasets are concatenated, randomly shuffled, and split into training and test sets using a $0.8/0.2$ ratio.
Optimization is performed using mini-batches of size $256$.

The discriminator is a convolutional network operating on single-channel inputs of shape $(1, 48, 36)$.
Its architecture follows \citet{habi2021raingan} and consists of three convolutional blocks.
Each block applies a $5\times5$ convolution with stride $2$ and padding $2$, followed by a LeakyReLU activation.
Channel dimensions increase as $1 \to d \to 2d \to 4d$.
The resulting feature maps are flattened and passed through a linear layer to produce a single logit.

Training minimizes the binary cross-entropy loss with logits using the Adam optimizer with a learning rate of $10^{-4}$ for $200$ epochs.
At evaluation time, predicted probabilities are thresholded at $0.5$ to obtain class labels.

\begin{figure}[t]
    \centering
    \begin{subfigure}[c]{0.495\textwidth}
        \centering%
        \includegraphics[width=1.0\linewidth]{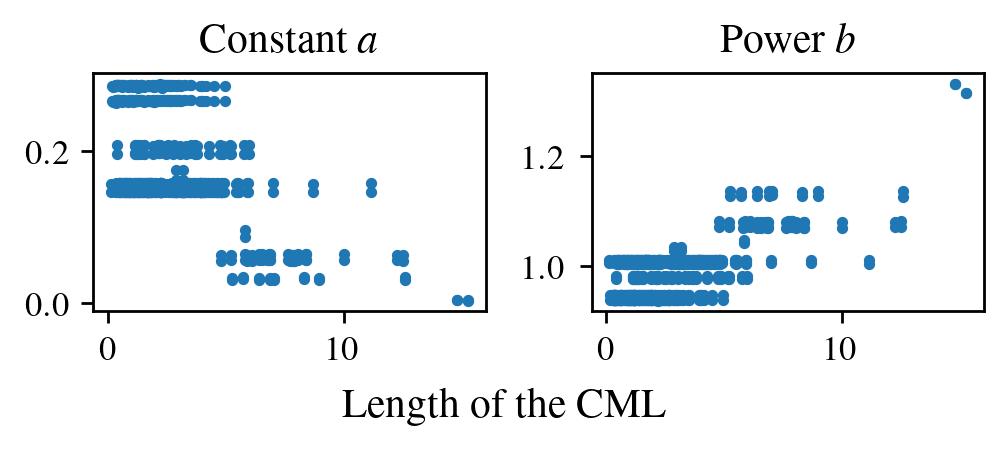}%
    \end{subfigure}%
    \hfill%
    \begin{subfigure}[c]{0.495\textwidth}
        \centering%
        \includegraphics[width=1.0\linewidth]{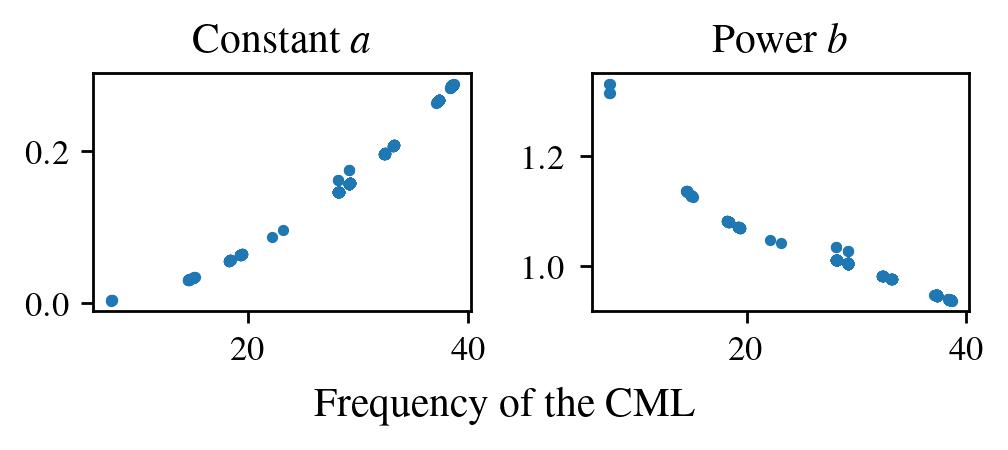}
    \end{subfigure}%
    \caption{Power-law parameters of the CML as a function of link's length and the frequency on the OpenMRG dataset.}
    \label{fig:cmls-power-law-stats}
\end{figure}

\section{Baselines Implementation}
\label{apdx:implem-details}

In this section, we provide implementation details for all baseline methods against which we compare our approach. For the meteorological baselines, hyperparameters are presented in \Cref{table:hyperparameters_meteo}. The details about the hyperparameters of DM posterior samplers are summarized in \Cref{table:hyperparameters_ps}.
We highlight that all baseline hyperparameters were tuned manually. 
For completeness, we also report the runtime required to generate one reconstruction for each baseline in \Cref{table:runtime_baselines}. The meteorological baselines are substantially faster than the diffusion baselines, for which the reported runtimes are normalized by batch size, since 10 samples are generated jointly and the total runtime is divided by 10 to obtain a per-reconstruction comparison.

\subsection{Meteorological baselines}

\paragraph{\idw.}
\label{ref:idw_implem}
We implement inverse-distance weighting (IDW) following \citet{eshel2021quantitative,shepard1968two}.
Each CML link observation is represented as a point measurement located at the link midpoint.
For every grid location $s$, we compute a normalized weighted average using $p=2$ and  $\epsilon = 10^{-6}$ for numerical stability.
A region-of-influence radius $\mathrm{roi}$ is set to $4$ in order to discard distant observations. 

\paragraph{\gmz.}

We implement the iterative GMZ reconstruction procedure following \citet{goldshtein2009idw-gmz,eshel2021quantitative}, that we adapt from \texttt{PyNNcml}~\citep{PyNNcml2026github}.
We first map each link measurement to the attenuation domain via the standard power law $A = a X^{b}$ \citep{itu2005specific}. Each link is discretized into a fixed number of uniformly spaced VRGs. We use $K=5$ virtual points per link and run the algorithm for $T=20$ iterations. 
At each iteration, an intermediate field is reconstructed using the same IDW operator as in ~\ref{ref:idw_implem}. The reconstructed field is bilinearly sampled at the virtual points, and link consistency is enforced by projecting the sampled values such that the mean attenuation along each link matches the original transformed measurement. 
 
\paragraph{\ok.}
We apply ordinary kriging algorithm \citep{matheron1963principles,cressie1993statistics} using \texttt{OrdinaryKriging} function from \texttt{PyKrige} package \citep{murphy2025pykrige}. We use the Exponential variogram model, with parameters inferred using PyKrige's default L1-norm-based variogram fitting.

\subsection{Diffusion posterior sampling baselines}  

\paragraph{\mgdm.}
We implement the \mgdm\ sampler proposed by \citet{janati2025mixture} in Algorithm $2$, by adapting the reference implementation\footnote{\url{https://github.com/badr-moufad/mgdm}}.
In particular, the code is adapted to the VE setting.
We sample the intermediate steps uniformly and use one Gibbs sampling set. 

\paragraph{\daps.}
We refer to \citet[Algo 1]{zhang2024daps} for the method used in the \daps\ sampler implemented based on the released code\footnote{\url{https://github.com/zhangbingliang2019/DAPS}}. We use the Langevin MCMC sampler for consistency and adapted to the VE setting.

\paragraph{\reddiff.}
We implement the \reddiff\ sampler as described in \citet[Algo 1]{mardani2024reddiff}. We modify the implementation provided in released repository\footnote{\url{https://github.com/NVlabs/RED-diff}} to the VE setting. 

\paragraph{\crepe.}
We follow the algorithm presented in \citet[Algo 1]{he2025crepe} combined with the reference code\footnote{\url{https://github.com/jiajunhe98/CREPE-Controlling-diffusion-with-REPlica-Exchange}} provided by the authors. We adapted the \crepe sampler to our use case of VE formulation.
Following the authors, we use the prior model as a proposal and use the \dps\ intermediate likelihoods (also referred to as intermediate rewards) to define the annealing path that bridge the Gaussian $p_T$
and the target $\pi_0$.

\paragraph{\dps.} 
We modify the released code\footnote{\url{https://github.com/DPS2022/diffusion-posterior-sampling}} that implements \citet[Algorithm 1]{chung2023dps} to operate within VE-setting.

\paragraph{\tds.}
We implement \citet[Algorithm~1]{wu2023tds} in the VE setting.
As a proposal distribution, we use \dps\ \citep[Algorithm~1]{chung2023dps}, which differs from the original proposal by normalizing the log-gradient of the intermediate likelihood.
For the variance of the intermediate likelihoods, we use the surrogate $\sigma_i^2 + (\tau \sigma_t)^2$, where $\tau$ is a hyperparameter set to $\tau=1$, as it yields the best empirical performance.

\paragraph{\mgps.}
We implement the \mgps\ \citep[Algo 1]{moufad2025mgps} based on the released code\footnote{\url{https://github.com/YazidJanati/mgps}} by adapting it to our VE setting.
We use $\eta=0.5$ for selecting the midpoint and we do not use warm start for the algorithm.

\begin{table}[t]
    \centering
    \captionsetup{font=small}
    \caption{The hyperparameters for each meteorological algorithm. The symbol ``--'' indicates that the corresponding hyperparameter is not applicable to the given method.}
    \resizebox{0.5\textwidth}{!}{
    \begin{tabular}{lccccc}
    \toprule
    \text{Algorithm} & {p} & \text{ROI} & \text{\# Points per link} & \text{\# Iterations} & \text{Variogram} \\
    \midrule
    \idw & 2 & 6 & 1 & -- & -- \\
    \gmz & 2 & 6 & 5 & 20 & -- \\
    \ok & -- & -- & -- & -- & \text{\texttt{exponential}}\\
    \bottomrule
    \end{tabular}
    }
    \label{table:hyperparameters_meteo}
\end{table}
\begin{table}[t]
    \centering
    \captionsetup{font=small}
    \caption{The hyperparameters for each posterior sampling algorithm.
    We use Karras scheduler with $\sigma_{\min} = 2 \times 10^{-3}$ and $\sigma_{\max} = 100$ for GP experiments and $\sigma_{\min} = 2 \times 10^{-3}$ and $\sigma_{\max} = 80$ for rain field experiments.}
    \resizebox{0.82\textwidth}{!}{
    \begin{tabular}{llcc}
    \toprule
    \text{Algorithm} & \text{Base hyperparameters} & \multicolumn{2}{c}{Tasks} \\
    \cmidrule(lr){3-4}
    & & \text{Gaussian Process} & \text{Real CML links} \\
    \midrule
    \daps & 
        \makecell[l]{$n_{\text{steps}}=100$ \\ $N_{\text{ode}}=5$ \\ $\text{\texttt{Min ratio}}=0.01$ \\ $\text{\texttt{MCMC sampler}}=\text{\texttt{Langevin}}$ \\ $\rho=7$} &
        \makecell{$\text{\texttt{MCMC steps}}=100$ \\ $\eta_0=5\times 10^{-4}$} &
        \makecell{$\text{\texttt{MCMC steps}}=50$ \\ $\eta_0=2\times 10^{-4}$} \\ 
    \midrule
    \dps & 
        \makecell[l]{$\rho=7$} &
        \makecell{$n_{\text{steps}}=320$ \\ $\gamma=4$} &
        \makecell{$n_{\text{steps}}=420$ \\ $\gamma=1$} \\
    \midrule
    \mgdm &
        \makecell[l]{$\rho=7$ \\ $\text{\texttt{lr}}=3\times 10^{-2}$ \\ $\text{\texttt{n\_gibbs\_repetitions}}=1$ \\ $\text{\texttt{n\_gradient\_repetitions}}=10$ } &
        \makecell{$n_{\text{steps}}=32$ \\ $\text{\texttt{n\_denoising\_steps}}=3$} &
        \makecell{$n_{\text{steps}}=50$ \\ $\text{\texttt{n\_denoising\_steps}}=5$} \\
    \midrule
    \mgps &
        \makecell[l]{$\rho=7$ \\ $\text{\texttt{lr}}=3\times 10^{-2}$ \\ $\text{\texttt{n\_gradient\_steps}}=10$} &
        \makecell{$n_{\text{steps}}=64$} &
        \makecell{$n_{\text{steps}}=32$} \\
    \midrule
    \reddiff &
        \makecell[l]{$n_{\text{steps}}=1000$ \\ $\rho=5$ \\ $\text{\texttt{obs\_weight}}=1$ \\ $\text{\texttt{grad\_term\_weight}}=1$} &
        \makecell{$\text{\texttt{lr}}=10^{-1}$} &
        \makecell{$\text{\texttt{lr}}=5\times 10^{-3}$} \\
    \midrule
    \tds &
        \makecell[l]{$\rho=7$} &
        \makecell{$n_{\text{steps}}=320$ \\ $\gamma=4$ \\ $\text{\texttt{n\_particles}}=10$} &
        \makecell{$n_{\text{steps}}=420$ \\ $\gamma=1$ \\ $\text{\texttt{n\_particles}}=4$} \\
    \bottomrule
    \end{tabular}
    }
    \label{table:hyperparameters_ps}
\end{table}

\section{Details on Siddon Algorithm}
\label{apdx:siddon-algo}
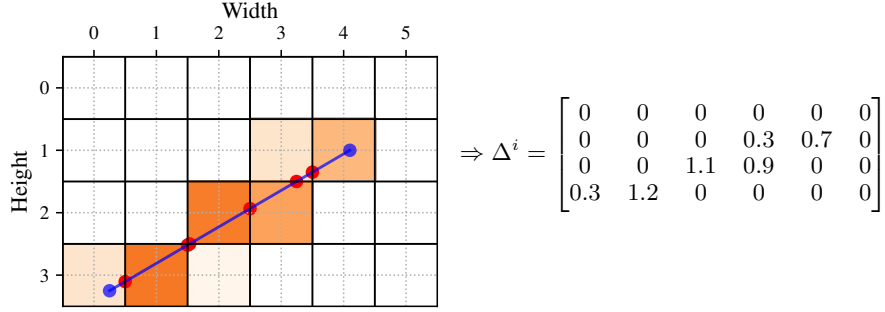
\begin{figure}[ht]
    \centering
    \begin{subfigure}[c]{0.35\textwidth}
        \centering%
        \includegraphics[width=\linewidth]{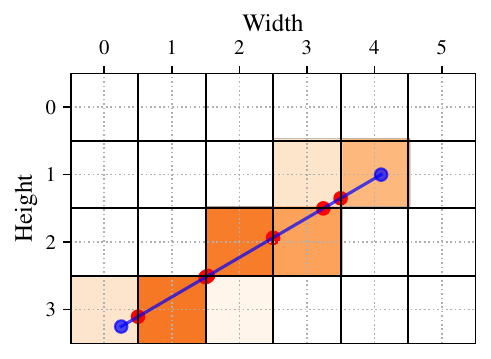}%
    \end{subfigure}%
    \begin{subfigure}[c]{0.24\textwidth}
        \centering%

\begin{tikzpicture}%
    \node[font=\small] {%
        $\Rightarrow \Delta^i = 
        \begin{bmatrix}
            0 & 0 & 0 & 0 & 0 & 0 \\
            0 & 0 & 0 & 0.3 & 0.7 & 0 \\
            0 & 0 & 1.1 & 0.9 & 0 & 0 \\
            0.3 & 1.2 & 0 & 0 & 0 & 0 \\
        \end{bmatrix}$
    };%
\end{tikzpicture}%
    \end{subfigure}%
    \caption{Illustration of a line sensor on a $6 \times 4$ grid.
    The {\color{blue}{blue dots}} define the line sensor whereas the {\color{red}{red dots}} mark its intersections with the two-dimensional grid.
    The color intensity of each cell is proportional to the intersection length: the darker, the higher.}
    \label{fig:cml-operator}
\end{figure}






We provide implementation details for computing the intersection-length weights $\Delta^i$ in \Cref{eq:weighted-sum-siddon}, i.e., the lengths of intersection between the $i$-th CML path and the $H\times W$ grid cells.

\paragraph{Grid Parametrization.}
Consider a regularly spaced grid with vertical and horizontal grid lines located at
\[
x=x_\mathrm{ref}+c\,\Delta x,\qquad y=y_\mathrm{ref}+r\,\Delta y,\qquad c,r\in\mathbb{Z},
\]
where $\Delta x,\Delta y>0$ are the grid spacings and $(x_\mathrm{ref},y_\mathrm{ref})$ specifies the grid origin.
Under the affine change of variables $\tilde x:=(x-x_\mathrm{ref})/\Delta x$ and $\tilde y:=(y-y_\mathrm{ref})/\Delta y$, these grid lines map to integers defined by $(\tilde x,\tilde y)$.
Hence, intersection computations can be carried out in the normalized coordinates and mapped back to the original coordinate system.

\emph{Considered convention.}\quad
In our setting, the radar maps use the origin $(-\tfrac{1}{2},-\tfrac{1}{2})$ and unit spacing in both directions, i.e., $\Delta x=\Delta y=1$.
Indexing cells by the integer coordinates of their centers $(c,r)\in\mathbb{Z}^2$ places the cell boundaries at $c\pm\tfrac{1}{2}$ and $r\pm\tfrac{1}{2}$; equivalently, cell centers lie at integer coordinates.

Therefore, the grid lines of the considered $H\times W$ discretization are located at
\[
x = c+\tfrac{1}{2},\qquad y = r+\tfrac{1}{2},
\quad
c\in\{-1,\dots,W-1\},\ \text{and}\ r\in\{-1,\dots,H-1\}.
\]

\paragraph{Line Parameterization.}
For a given link, let $\mathbf{s}_0,\mathbf{s}_1\in\mathbb{R}^2$ denote its start and end points, and define the affine interpolation
\[
\mathbf{s}_t \;:=\; \mathbf{s}_0 + t(\mathbf{s}_1-\mathbf{s}_0), \qquad t\in[0,1].
\]
We denote coordinates by $[\mathbf{s}_t]_x$ and $[\mathbf{s}_t]_y$; in particular, $[\mathbf{s}_0]_x,[\mathbf{s}_0]_y$ and $[\mathbf{s}_1]_x,[\mathbf{s}_1]_y$ for the start and end points respectively.

\paragraph{Algorithm.}
Let $\mathbf{d}:=\mathbf{s}_1-\mathbf{s}_0$ with $d_x:=[\mathbf{d}]_x$ and $d_y:=[\mathbf{d}]_y$.
We seek parameters $t\in(0,1)$ such that $\mathbf{s}_t$ lies on a grid line.
For each vertical grid line $x=c+\tfrac{1}{2}$ with $d_x\neq 0$, the corresponding parameter is
\[
t^x_c \;:=\; \frac{\big(c+\tfrac{1}{2}\big)-[\mathbf{s}_0]_x}{d_x}
\]
and we retain those satisfying $t^x_k\in(0,1)$.
Analogously, for each horizontal grid line $y=r+\tfrac{1}{2}$ with $d_y\neq 0$,
\[
t^y_r \;:=\; \frac{\big(r+\tfrac{1}{2}\big)-[\mathbf{s}_0]_y}{d_y},
\]
retaining those with $t^y_c\in(0,1)$.
We then form
\[
T \;:=\; \{0,1\}\;\cup\;\{t^x_c:\ t^x_c\in(0,1)\}\;\cup\;\{t^y_r:\ t^y_r\in(0,1)\},
\]
where the set union removes duplicates (e.g., when the line crosses a grid corner).
Sorting $T$ yields
\[
0=t_0<t_1<\cdots<t_N=1.
\]
Consecutive parameters define sub-segments that lie within single grid cells. Their Euclidean lengths are
\[
\delta_n \;:=\;\|\mathbf{s}_{t_{n+1}}-\mathbf{s}_{t_n}\|_2
\;=\;\|\mathbf{s}_1-\mathbf{s}_0\|_2\,(t_{n+1}-t_n),\qquad n=0,\dots,N-1.
\]
We store the weights as a sparse matrix $\Delta^{i}\in\mathbb{R}^{H\times W}_{\ge 0}$, where $[\Delta^{i}]_{r,c}$ equals the length of the portion of the link in cell $(c,r)$.
Vectorizing $\Delta^i$ yields the coefficients $\{\Delta^i_k\}_{k=1}^{HW}$ used in \Cref{eq:weighted-sum-siddon}.
An example of the resulting sparse weight matrix is shown in \Cref{fig:cml-operator}.

\section{Closed-form Expression of the Oracle Posterior}
\label{apdx:gp-exp}

\paragraph{Setting.}
We recall that the considered setting: a linear inverse problem where the prior $X$ is a GP $X\sim\mathcal{GP}(0,k)$ on $[-5,5]$, with the RBF kernel
\begin{equation}
k(s,s')=\exp\!\left(-\frac{(s-s')^2}{2\ell^2}\right),
\qquad \ell=0.6.
\end{equation}
The inverse problem is defined by
\begin{equation}
Y_i=\int_{a_i}^{b_i}X(s) \, \rmd s + \sigma Z_i,
\qquad Z_i\sim\mathcal{N}(0,1),
\end{equation}
where $\{[a_i,b_i]\}_{i=1}^m$ are the considered integration intervals and $\{Z_i\}_{i=1}^m$ are mutually independent and independent of $X$.
Since the inverse problem involves linear transformations of the GP prior,
the posterior is also a GP with closed-form mean and covariance \citep{rasmussen2006gp} that writes
\begin{align*}
\mu_{X|y}(s)&=\mathbf{k}_y(s)^\top(\mathbf{K}_{yy}+\sigma^2\mathbf{I})^{-1}\mathbf{y}, \\
k_{X|y}(s,s')&=k(s,s')-\mathbf{k}_y(s)^\top(\mathbf{K}_{yy}+\sigma^2\mathbf{I})^{-1}\mathbf{k}_y(s'),
\end{align*}
where $\mathbf{k}_y(s) \in \mathbb{R}^m$ is the covariance between $X(s)$ and the $i$-th observation $y_i$, namely $\mathrm{cov}(X(s), y_i)$; and $\mathbf{K}_{yy} \in \mathbb{R}^{m \times m}$ covariance between the observations $[y_1, \ldots, y_m]$, namely $[\mathbf{K}_{yy}]_{ij}$ is the covariance between the $i$-th and $j$-th observations $\mathrm{cov}(y_i, y_j)$.
Getting the closed-form expression of the posterior GP amounts to deriving the expressions of $[\mathbf{k}_y(s)]_i$ and $[\mathbf{K}_{yy}]_{ij}$ for all $i, j$ in $\intset{1}{m}$.

\paragraph{Derivation of the covariances.}
The covariance between the $i$-th observation and $X(s)$ writes as
\begin{align*}
    [\mathbf{k}_y(s)]_i 
        & = \mathrm{cov}\left(X(s), \int_{a_i}^{b_i} X(s') \rmd s' + \sigma Z_i \right) \\
        & = \int_{a_i}^{b_i} \mathrm{cov}\big(X(s), X(s')\big) \rmd s' \\
        & = \int_{a_i}^{b_i} k(s, s') \rmd s'
\end{align*}
where the second equality follows from the linearity of the integral and the independence between $X$ and $Z_i$.
The value $[\mathbf{k}_y(s)]_i$ can be numerically approximated using the Error Function $\mathrm{erf}$\footnote{\url{https://docs.pytorch.org/docs/stable/generated/torch.erf.html}}
\begin{equation}
    \label{eq:rbf-int-1}
    [\mathbf{k}_y(s)]_i 
        = \ell \sqrt{\frac{\pi}{2}} \left[ \mathrm{erf}\left( \frac{b_i - s}{\sqrt{2}\ell} \right) - \mathrm{erf}\left( \frac{a_i - s}{\sqrt{2}\ell} \right) \right].
\end{equation}
Similarly, the covariance between the $i$-th and $j$-th observations writes
\begin{align*}
    [\mathbf{K}_{yy}]_{ij}
        & = \mathrm{cov}\left(\int_{a_i}^{b_i} X(s) \rmd s + \sigma Z_i , \int_{a_j}^{b_j} X(s') \rmd s' + \sigma Z_j \right) \\
        & = \int_{a_i}^{b_i} \int_{a_j}^{b_j}  \mathrm{cov}\big(X(s), X(s')\big) \rmd s \rmd s' + \sigma^2 \delta_{i=j} \\
        & = \int_{a_i}^{b_i} \int_{a_j}^{b_j}  k(s,s') \rmd s \rmd s' \  +  \ \sigma^2 \delta_{i=j},
\end{align*}
where $\delta_{i=j}$ equals $1$ if $i=j$ and zero otherwise.
The first term on the right-hand side can be computed by: (i) using the result in \Cref{eq:rbf-int-1}, and (ii) integrating by leveraging the formula of the primitive of $s \mapsto \mathrm{erf}({s}/{\sqrt{2\ell}})$, namely
$$
H(z) = z \cdot \text{erf}\left( \frac{z}{\sqrt{2}\ell} \right) + \sqrt{\frac{2}{\pi}} \ell \exp\left( -\frac{z^2}{2\ell^2} \right).
$$
Putting it together, the expression of $[\mathbf{K}_{yy}]_{ij}$ writes as
$$
[\mathbf{K}_{yy}]_{ij} = \ell \sqrt{\frac{\pi}{2}} \Big[ H(b_i - a_j) - H(a_i - a_j) - H(b_i - b_j) + H(a_i - b_j) \Big]
 +  \ \sigma^2 \delta_{i=j}.
$$

\begin{figure*}[!t]
    \centering 
    \includegraphics[width=\textwidth]{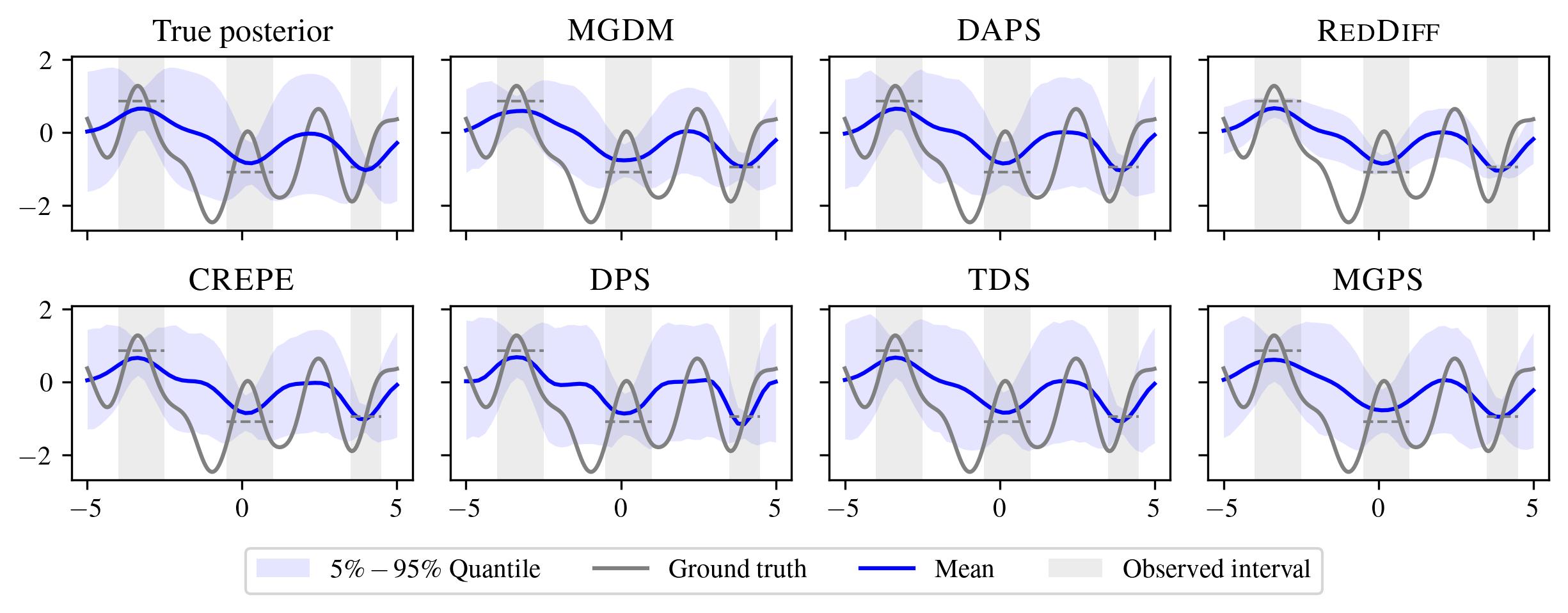}
    \captionsetup{font=small}
    \caption{Extended results of \Cref{fig:gp-exp-viz-1}. Comparison between the reconstructions of the baselines on the inverse problem with diffusion prior on the setting of GP.
    The y-axis shows intensities while the x-axis represents a one-dimensional grid of the $[-5, 5]$ interval.
    The dashed horizontal lines depict values of the integral over the observed intervals.}
    \label{fig:gp-exp-viz-2}
\end{figure*}


\section{Censored Gaussian Process}
\label{apdx:censored-gp}

\paragraph{Setting revisted.}
We revisit the setting in \Cref{sec:censored-gp}.
We assume that the rain field $X:\Omega\subset\mathbb{R}^2\rightarrow\mathbb{R}_+$ is described by a left-censored GP with a power transformation $\beta>1$
$$
X(\mathbf{s}) = \max\left(0, V(\mathbf{s})^\beta\right), \qquad \mathbf{s} \in \Omega \subset \mathbb R^2,
$$
namely, $V\sim\mathcal{GP}(\mu,k)$ being a stationary GP with constant $\mu\in\mathbb{R}$ and kernel
\begin{equation*}
    k(\mathbf{s}, \mathbf{s}') = \sigma^2 \exp \left( -\frac{1}{2} ( \mathbf{s} - \mathbf{s}' )^\top \mathbf{Q}^{-2} \ ( \mathbf{s} - \mathbf{s}' ) \right),
    \quad \sigma > 0,
    \quad \mathbf{Q} = \diag{{\ell}_{1}, {\ell}_{2}} \in \mathbb{R}^{2\times2}
    ,
\end{equation*}
where the positive definite matrix $\mathbf{Q}$ is meant to account for anisotropy of the field.
The domain of interest $\Omega\subset\mathbb{R}^2$ is discretized in a grid of $H\times W$ at the points $\{s_k\}_{k=1}^{HW}$ and we assume to have access to the realization of $N$ of the discretized field $x_i \in \mathbb{R}^{HW}_{+}$, that is, for $i \in \intset{1}{N}$, we have
$$
x_i = \max(0, v_i^\beta), 
\quad \text{ where } \quad v_i \sim \mathcal{N}\left(\mu \mathbf 1, \Gamma(\mathbf{Q})\right)
\quad \text{ and } \quad \Gamma_{\ell,j} = k(\mathbf{s}_\ell, \mathbf{s}_j), \; \text{ for } 0 \leq \ell, j \leq HW
$$
The goal is to estimate the paramters of the latent model $(\mu, \mathbf{Q}, \sigma, \beta)$. To do so, we use the EM algorithm and proceed first by deriving the algorithm when $\beta=1$, and then generalize to $\beta > 1$.

\paragraph{Parameters estimation of the model.}
One way to estimate the parameters and avoid optimizing jointly $\beta$, is to fit the model parameters for different values of $\beta$ and then select the configuration that gives the best classifier two-sample test.

\subsection{EM algorithm for $\beta = 1$}

The joint model factorizes as
$$
p(x_i, v_i) = \delta_{\max(0, v_i)}(x_i)\, p_V(v_i),
$$
where $p_V$ denotes the marginal distribution of $v_i$ and is given by the Gaussian
$\mathcal{N}(\mu \mathbf{1}, \Gamma(\mathbf{Q}))$.
For a given observation $x_i$, we define the index sets corresponding to positive and zero components as
$$
I_i^+ = \{k : [x_i]_k > 0\}, \qquad I_i^0 = \{k : [x_i]_k = 0\},
$$
and denote by $(x_i^+, x_i^0)$ the restriction of $x_i$ to $I_i^+$ and $I_i^0$, respectively.

\emph{E-step.}\quad
The E-step consists in sampling from the posterior $p_{V|X}(\cdot \mid x_i)$, which factorizes as
\begin{equation}
\label{eq:post-censored-gp}
p_{V|X}(v_i \mid x_i)
\propto \delta_{x_i^+}(v_i^+) \times \mathbf{1}_{\{v_i^0 \le 0\}}\, p_V(v_i^0 \mid v_i^+ = x_i^+).
\end{equation}
The second factor corresponds to a truncated multivariate Gaussian distribution.
Samples are obtained via Gibbs sampling, leveraging the fact that one-dimensional truncated Gaussian conditionals can be sampled exactly using inverse transform sampling.

\emph{M-step.}\quad
At iteration $k$, posterior samples $v_i \sim p_{V|X}^{(k)}(\cdot \mid x_i)$ are drawn using the current parameter estimates.
The parameters $(\mu, Q)$ are then updated by maximizing the complete-data log-likelihood,
$$
\arg\max_{\mu, \mathbf{Q}, \sigma} \sum_{i=1}^N \log p_V(v_i),
$$
which corresponds to standard maximum likelihood estimation for a multivariate Gaussian model.

\subsection{EM algorithm for $\beta > 1$}

For $\beta > 1$, the posterior distribution $p_{V|X}^\beta(\cdot \mid x_i)$ no longer involves a truncated multivariate Gaussian.
Nevertheless, its density admits a closed-form expression:
\begin{equation}
\label{eq:density-censored-gp-beta}
p_V^\beta(v_i^0 \mid v_i^+ = x_i^+)
= p_V\!\left((v_i^0)^{1/\beta} \mid v_i^+ = x_i^+\right)
\Big| \det\!\left(\mathrm{J}_f(v_i^0)\right) \Big|, 
\end{equation}
where $\mathrm{J}_f(v_i^0)$ denotes the Jacobian of the elementwise power transformation $f(v) = v^\beta$.
The Jacobian is diagonal and can be computed exactly and efficiently.

\emph{E-step.}\quad
The E-step is performed using a Metropolis-within-Gibbs scheme.
We use the same procedure in E-step of the case $\beta=1$ as a proposal distribution, and acceptance ratios are computed using the density in \eqref{eq:density-censored-gp-beta}.

\emph{M-step.}\quad
Although the power transformation modifies the posterior density, it does not affect the model parameters.
Consequently, the M-step mirrors that of the $\beta=1$ case, with the exception that posterior samples are transformed back by applying the inverse power.
Specifically, parameter updates are obtained by maximizing
$$
\arg\max_{\mu, \mathbf{Q}, \sigma} \sum_{i=1}^N \log p_V\!\left(v_i^{1/\beta}\right).
$$
which, similarly, correspond maximum likelihood estimation for a multivariate Gaussian model expect the inputs are $v_i^{1/\beta}$.

\subsection{Practical Computational Challenges}

While the procedure described above enables learning from censored processes, it scales poorly in our current high-dimensional setting, where rain fields have dimension $48 \times 36 = 1728$ and nearly $50\%$ of the entries are censored.
The E-step requires posterior sampling to impute the censored coordinates of each rain field by sampling from a truncated multivariate Gaussian distribution. Since the censored coordinates vary across rain fields, this limits the ability to leverage vectorized computation and makes estimation prohibitively dependent on the number of rain fields, which is $21{,}196$ in our setting.
In addition, using Gibbs sampling for the truncated multivariate Gaussian requires a long burn-in period due to the dimensionality of the data.

\section{Examples of Reconstructions in CML Experiments}

We present qualitative examples of rain field reconstructions from CML measurements on the OpenMRG dataset.
The comparison includes all considered baselines, both meteorological and diffusion-based methods.
Three representative reconstruction examples are shown.

\begin{figure}[t]
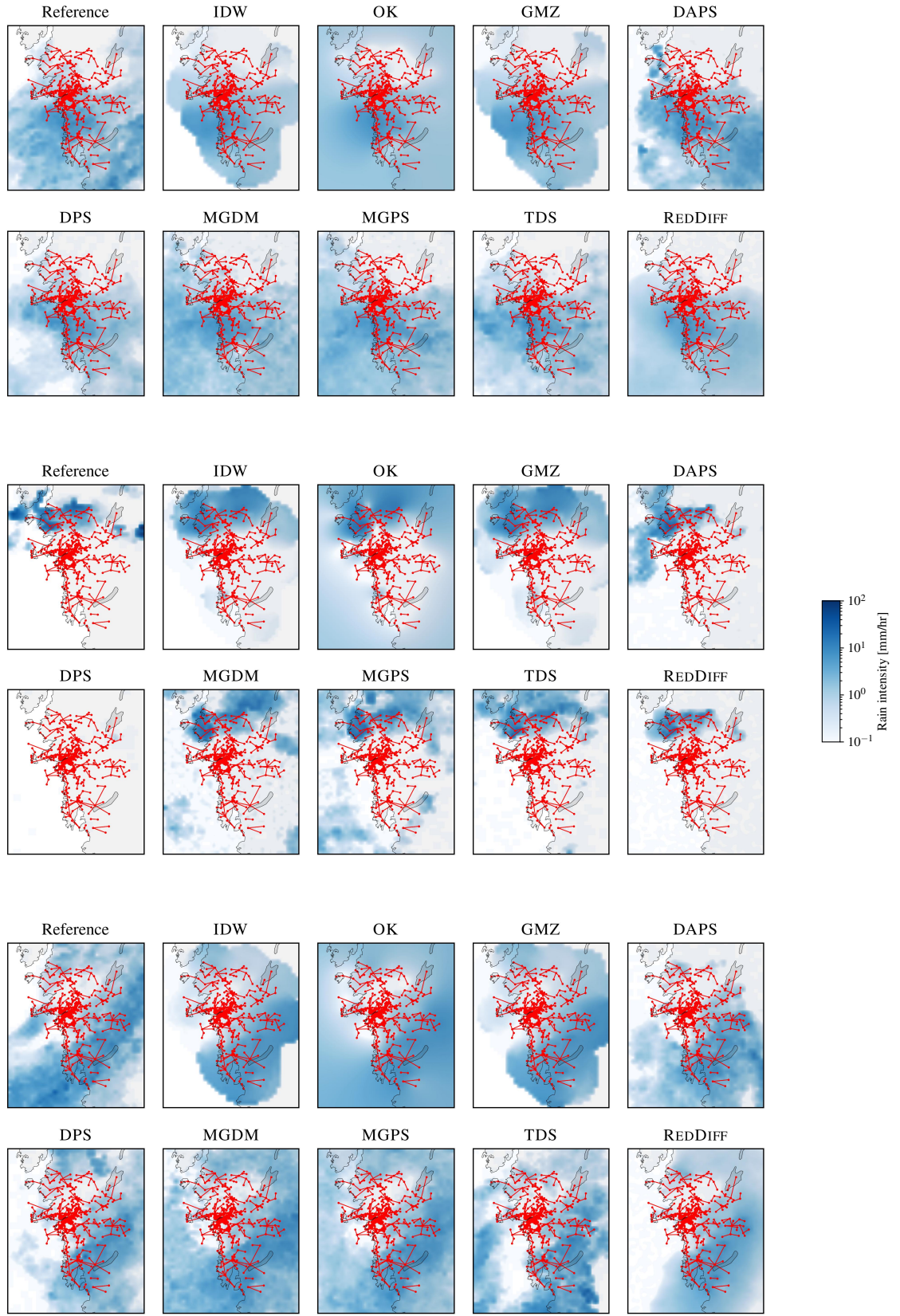

    \foreach \fieldidx in {3403,3156,3390}{%
        \centering%
        \foreach \method in {ref,idw,ok,gmz,daps}{%
            \begin{subfigure}[t]{0.15\textwidth}
                \centering
                \includegraphics[width=\textwidth]{files/figs/rain_recs_real_jpeg/\fieldidx/\method.jpeg}
            \end{subfigure}%
        }
            
        \centering%
        \foreach \method in {dps,mgdm,mgps,tds,reddiff}{%
            \begin{subfigure}[t]{0.15\textwidth}
                \centering
                \includegraphics[width=\textwidth]{files/figs/rain_recs_real_jpeg/\fieldidx/\method.jpeg}
            \end{subfigure}%
        }

        \vspace*{8mm}
    }
    \hfill\begin{subfigure}[t]{0.08\textwidth}
        \centering
        \vspace{-134mm} 
        \includegraphics[width=1.\textwidth]{files/figs/rain_color_bar.pdf}
    \end{subfigure}%
    \captionsetup{font=small}
    \vspace*{-8mm} 
    \caption{Qualitative comparisons of rain field reconstructions on real CMLs links from OpenMRG dataset on three reconstruction tasks. The network of CMLs is depicted in red.}
    \label{fig:real-cmls-exp-all-comp-1}
\end{figure}

\begin{figure}[t]
    \foreach \fieldidx in {4654,3096,3106}{%
        \centering%
        \foreach \method in {ref,idw,ok,gmz,daps}{%
            \begin{subfigure}[t]{0.15\textwidth}
                \centering
                \includegraphics[width=\textwidth]{files/figs/rain_recs_real_jpeg/\fieldidx/\method.jpeg}
            \end{subfigure}%
        }
            
        \centering%
        \foreach \method in {dps,mgdm,mgps,tds,reddiff}{%
            \begin{subfigure}[t]{0.15\textwidth}
                \centering
                \includegraphics[width=\textwidth]{files/figs/rain_recs_real_jpeg/\fieldidx/\method.jpeg}
            \end{subfigure}%
        }

        \vspace*{8mm}
    }
    \hfill\begin{subfigure}[t]{0.08\textwidth}
        \centering
        \vspace{-134mm} 
        \includegraphics[width=1.\textwidth]{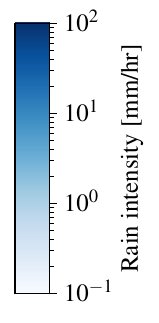}
    \end{subfigure}%
    \captionsetup{font=small}
    \vspace*{-8mm} 
    \caption{Another set of qualitative comparisons of rain field reconstructions on real CMLs links from OpenMRG dataset on three reconstruction tasks. The network of CMLs is depicted in red.}
    \label{fig:real-cmls-exp-all-comp-2}
\end{figure}




\end{document}